%% file: arxiv.tex
\documentclass[10pt,twocolumn,letterpaper]{article}

\usepackage[pagenumbers]{cvpr} %

\input{packages}

\input{preamble}

\definecolor{cvprblue}{rgb}{0.21,0.49,0.74}
\usepackage[pagebackref,breaklinks,colorlinks,allcolors=cvprblue]{hyperref}
\usepackage[capitalize]{cleveref}
\usepackage{graphicx}
\usepackage{multirow}
\usepackage{array}
\usepackage{tikz}       %
\usepackage{pgfplots}   %

\crefname{section}{Sec.}{Secs.}
\Crefname{section}{Section}{Sections}
\Crefname{table}{Table}{Tables}
\crefname{table}{Tab.}{Tabs.}
\Crefname{appendix}{Appendix}{Appendixes}
\crefname{appendix}{Appx.}{Appxs.}

\definecolor{darkred}{rgb}{0.7,0.1,0.1}
\definecolor{darkgreen}{rgb}{0.1,0.6,0.1}
\definecolor{cyan}{rgb}{0.7,0.0,0.7}
\definecolor{otherblue}{rgb}{0.1,0.4,0.8}
\definecolor{maroon}{rgb}{0.76,.13,.28}
\definecolor{burntorange}{rgb}{0.81,.33,0}

\def\paperID{18772} 

\def\confName{CVPR}
\def\confYear{2025}

\title{Memories of Forgotten Concepts}

\author{
        Matan Rusanovsky$^1$\footnotemark[1] \hspace{5mm}
        Shimon Malnick$^1$\footnotemark[1]\hspace{5mm}
        Amir Jevnisek$^1$\footnotemark[1]\hspace{5mm}
        Ohad Fried$^2$ \hspace{5mm}
       Shai Avidan$^1$
        \\[14pt]
        $^1$Tel Aviv University \hspace{10mm} $^2$Reichman University
        \\
        \small{\url{https://matanr.github.io/Memories_of_Forgotten_Concepts}}
        \\
}

\begin{document}

\maketitle

\renewcommand\thefootnote{*}
\footnotetext{Equal contribution.}
\renewcommand\thefootnote{\arabic{footnote}} %

\def\ArxivVersion{}

\input{macros}

\input{sections/00_abstract}  
\input{sections/01_intoduction}

\input{sections/02_related_work}

\input{sections/03_analysis}

\input{sections/06_limitations}
\input{sections/08_conclusions}

\vspace{-7pt}
\paragraph{Acknowledgments.} This project was supported by ISF grants No.\ 1574/21 and 2132/23.

{
    \small
    \bibliographystyle{ieeenat_fullname}
    \bibliography{main}
}

\input{appendix}

\end{document}

%% file: packages.tex
\usepackage{times}
\usepackage{epsfig}
\usepackage{graphicx}
\usepackage{amsmath}
\usepackage{amssymb}
\usepackage{booktabs}
\usepackage{algorithm}
\usepackage[noend]{algpseudocode}
\usepackage{csquotes}
\usepackage{svg}
\usepackage{dcolumn}
\usepackage{tikz}
\usepackage[dvipsnames]{xcolor}
\usepackage{pifont}
\usetikzlibrary{positioning,calc,arrows,shapes,decorations.text}
\usepackage{mathtools}
\usepackage{multirow}
\usepackage{makecell}
\usepackage{bm}
\usepackage{cancel}
\usepackage{caption}
\usepackage{subcaption}

\usepackage[export]{adjustbox}
\usepackage[hyphens]{url}
\usepackage[normalem]{ulem}

\useunder{\uline}{\ul}{}

%% file: preamble.tex
\DeclareMathOperator{\nllzt}{NLL_{\rightarrow z_T}}

%% file: macros.tex
\newcommand{\betweencellpdf}{\ensuremath{h^c}}
\newcommand{\betweencellpdffine}{\ensuremath{h^f}}
\newcommand{\approxbetweencellpdffine}{\ensuremath{\hat{h}^f}}
\newcommand{\incellpdf}{\ensuremath{f}}
\newcommand{\incellpdfnormalized}{\ensuremath{f'}}

\newcommand{\incellcdf}{\ensuremath{F}}

\newcommand{\totalpdf}{\ensuremath{f_{dd}}}
\newcommand{\totalcdf}{\ensuremath{F_{dd}}}

\newcommand{\ignorethis}[1]{}
\newcommand{\redund}[1]{#1}

\newcommand{\aposteriori}     {\textit{a~posteriori}}
\newcommand{\perse      }     {\textit{per~se}}
\newcommand{\naive      }     {{na\"{\i}ve}}
\newcommand{\Naive      }     {{Na\"{\i}ve}}
\newcommand{\Identity   }     {\mat{I}}
\newcommand{\Zero       }     {\mathbf{0}}
\newcommand{\Reals      }     {{\textrm{I\kern-0.18em R}}}
\newcommand{\isdefined  }     {\mbox{\hspace{0.5ex}:=\hspace{0.5ex}}}
\newcommand{\texthalf   }     {\ensuremath{\textstyle\frac{1}{2}}}
\newcommand{\half       }     {\ensuremath{\frac{1}{2}}}
\newcommand{\third      }     {\ensuremath{\frac{1}{3}}}
\newcommand{\fourth     }     {\ensuremath{\frac{1}{4}}}

\newcommand{\Lone} {\ensuremath{L_1}}
\newcommand{\Ltwo} {\ensuremath{L_2}}

\newcommand{\degree} {\ensuremath{^{\circ}}}

\newcommand{\mat        } [1] {{\text{\boldmath $\mathbit{#1}$}}}
\newcommand{\Approx     } [1] {\widetilde{#1}}
\newcommand{\change     } [1] {\mbox{{\footnotesize $\Delta$} \kern-3pt}#1}

\newcommand{\Order      } [1] {O(#1)}
\newcommand{\set        } [1] {{\lbrace #1 \rbrace}}
\newcommand{\floor      } [1] {{\lfloor #1 \rfloor}}
\newcommand{\ceil       } [1] {{\lceil  #1 \rceil }}
\newcommand{\inverse    } [1] {{#1}^{-1}}
\newcommand{\transpose  } [1] {{#1}^\mathrm{T}}
\newcommand{\invtransp  } [1] {{#1}^{-\mathrm{T}}}
\newcommand{\relu       } [1] {{\lbrack #1 \rbrack_+}}

\newcommand{\abs        } [1] {{| #1 |}}
\newcommand{\Abs        } [1] {{\left| #1 \right|}}
\newcommand{\norm       } [1] {{\| #1 \|}}
\newcommand{\Norm       } [1] {{\left\| #1 \right\|}}
\newcommand{\pnorm      } [2] {\norm{#1}_{#2}}
\newcommand{\Pnorm      } [2] {\Norm{#1}_{#2}}
\newcommand{\inner      } [2] {{\langle {#1} \, | \, {#2} \rangle}}
\newcommand{\Inner      } [2] {{\left\langle \begin{array}{@{}c|c@{}}
                               \displaystyle {#1} & \displaystyle {#2}
                               \end{array} \right\rangle}}

\newcommand{\twopartdef}[4]
{
  \left\{
  \begin{array}{ll}
    #1 & \mbox{if } #2 \\
    #3 & \mbox{if } #4
  \end{array}
  \right.
}

\newcommand{\fourpartdef}[8]
{
  \left\{
  \begin{array}{ll}
    #1 & \mbox{if } #2 \\
    #3 & \mbox{if } #4 \\
    #5 & \mbox{if } #6 \\
    #7 & \mbox{if } #8
  \end{array}
  \right.
}

\newcommand{\len}[1]{\text{len}(#1)}

\newlength{\w}
\newlength{\h}
\newlength{\x}

\definecolor{darkred}{rgb}{0.7,0.1,0.1}
\definecolor{darkgreen}{rgb}{0.1,0.6,0.1}
\definecolor{cyan}{rgb}{0.7,0.0,0.7}
\definecolor{otherblue}{rgb}{0.1,0.4,0.8}
\definecolor{maroon}{rgb}{0.76,.13,.28}
\definecolor{burntorange}{rgb}{0.81,.33,0}
\definecolor{cyan_real}{rgb}{0.0,1.0,1.0}
\definecolor{purple_real}{rgb}{0.5,0.0,0.5}

\ifdefined\ShowNotes
  \newcommand{\colornote}[3]{{\color{#1}\textbf{#2} #3\normalfont}}
\else
  \newcommand{\colornote}[3]{}
\fi

\newcommand {\todo}[1]{\colornote{cyan}{TODO}{#1}}
\newcommand {\ohad}[1]{\colornote{burntorange}{OF:}{#1}}
\newcommand {\shai}[1]{\colornote{darkgreen}{SA:}{#1}}
\newcommand {\shimon}[1]{\colornote{otherblue}{SM:}{#1}}
\newcommand {\matan}[1]{\colornote{cyan}{MR:}{#1}}
\newcommand {\amir}[1]{\colornote{maroon}{AJ:}{#1}}

\newcommand {\reqs}[1]{\colornote{red}{\tiny #1}}

\newcommand {\new}[1]{\colornote{red}{#1}}

\newcommand*\rot[1]{\rotatebox{90}{#1}}

\newcommand {\newstuff}[1]{#1}

\newcommand\todosilent[1]{}

\newcommand{\woBGmask}{{w/o~bg~\&~mask}}
\newcommand{\woMask}{{w/o~mask}}

\providecommand{\keywords}[1]
{
  \textbf{\textit{Keywords---}} #1
}

\newcommand{\R}{\mathbb{R}}
\newcommand{\D}{\mathcal{D}}
\newcommand{\N}{\mathcal{N}}
\newcommand{\DC}{\D_\mathcal{C}}
\newcommand{\DT}{\D_\mathcal{T}}
\newcommand{\X}{\mathcal{X}}
\newcommand{\Z}{\mathcal{Z}}

\newcommand{\probP}{\text{I\kern-0.15em P}}

\newcommand{\argmin}{\mathop{\mathrm{argmin}}}
\newcommand{\argmax}{\mathop{\mathrm{argmax}}}
\newcommand{\pr}[1]{\left(#1\right)}

\ifdefined\ArxivVersion
  \newcommand{\arxiv}[2]{#1}
\else
  \newcommand{\arxiv}[2]{#2}
\fi

\ifdefined\RevisedVersion
  \newcommand{\revision}[1]{{\color{darkred}{#1}}}
\else
  \newcommand{\revision}[1]{#1}
\fi

%% file: sections/00_abstract.tex
\begin{abstract}
Diffusion models dominate the space of text-to-image generation, yet they may produce undesirable outputs, including explicit content or private data. 
To mitigate this, concept ablation techniques have been explored to limit the generation of certain concepts.
In this paper, we reveal that the erased concept information persists in the model and that erased concept images can be generated using the right latent. 
Utilizing inversion methods, we show that there exist latent seeds capable of generating high quality images of erased concepts. %
Moreover, we show that these latents have likelihoods that overlap with those of images outside the erased concept.
We extend this to demonstrate that for every image from the erased concept set, we can generate many seeds that generate the erased concept.
Given the vast space of latents capable of generating ablated concept images, our results suggest that fully erasing concept information may be intractable, highlighting possible vulnerabilities in current concept ablation techniques.

\end{abstract}

%% file: sections/01_intoduction.tex
\input{figures/25_teaser/figure}
\section{Introduction}
\label{sec:intro}

Diffusion models have emerged as a prominent tool for text-to-image tasks, extending their importance beyond the research community. Researchers have developed methods to utilize diffusion models for text-guided image editing, increasing their popularity even further. However, it has been demonstrated~\cite{schramowski2023sld} that these models can generate undesirable content, such as violent and explicit material. This highlights the importance of developing techniques to ablate (\ie, forget or erase) specific concepts (\eg, objects, styles).

A plethora of studies have focused on erasing concepts from diffusion models. Erased concepts are described by text, and the weights of the model are steered away from generating images that are associated with these texts. 
Then, the ablated model is expected to generate images that do not belong to the population of the erased concept, when introduced with the text describing the erased concept. But, does this mean the concept is erased? Can the model still generate images of the erased concept in some other way?

In our work, forgetting an image means that the ablated model can no longer generate a specific image (say, a specific image of a church) with a reasonable likelihood. A more interesting extension is forgetting a concept, which means that an ablated model can no longer produce images that are categorized as belonging to the ablated concept (say, the model can no longer produce any image containing any church) with a reasonable likelihood. 
Here we take memory to mean that the model can generate an image or a concept, regardless of whether that image was part of the training process, or part of the generalization capabilities of the model.

To date, the analysis of ablated models is mainly done on the output image, as 
shown in \cref{fig:teaser}(left). Given an ablated text (\ie, ``Starry Night") and a random seed, a model devoid of Van Gogh's style produces an image that is not in the style of Van Gogh. Analysis done on this image will confirm that this is indeed the case. In contrast, we test the following hypothesis:

{\em 
\textbf{Hypothesis:} An ablated model should not have a high likelihood seed vector that can be used to generate a high-quality ablated image.}

This paper deals with ways to analyze this hypothesis. In our analysis, we assume that both the ablated text prompt {\em and} the target ablated image are given. We then measure both the likelihood, in latent space, of the corresponding seed, and the quality of the generated image (\cref{fig:teaser}(right)).
For an effectively erased model, it should not be possible to identify a latent seed that is both likely and yields a high-quality ablated image. However, our analysis shows that the opposite holds true. Models that were ablated using state-of-the-art methods can still generate high-quality ablated images from high-likelihood seeds (\cref{fig:concept-level-results}).

Technically, to do that, we use {\em diffusion inversion} to find a latent seed vector that corresponds to the ablated image. We further analyze the identified seed vectors and find that they are as likely, in latent space, as seeds of normal (\ie, non-ablated) images. This suggests that ablated models {\em do not} forget ablated concepts. We also find that multiple, distinct, seed vectors can be used to generate an ablated image. We show that by using multiple random {\em support} images, it is possible to obtain seed vectors with high likelihood that can generate a given {\em query} image. These observations suggest that information about the ablated concept persists within the latent space, thereby questioning the effectiveness of concept ablation in these models. 

To summarize, we make the following contributions:
\begin{itemize}
    \item Introduce a metric to {\em analyze} how much an ablated model remembers erased concepts and images. We demonstrate it on $9$ recently published methods and $6$ different concepts.
    \item Show that diffusion inversion of ablated images recovers latent seed vectors with high likelihood and generates images with high PSNR score.
    \item Show that a single image can be inverted to multiple \emph{distant} seeds, suggesting that erasing is harder than it looks.
\end{itemize}

%% file: figures/25_teaser/figure.tex
\begin{figure*}[t]
    \centering
    \resizebox{1 \textwidth}{!}{
    \includegraphics[]{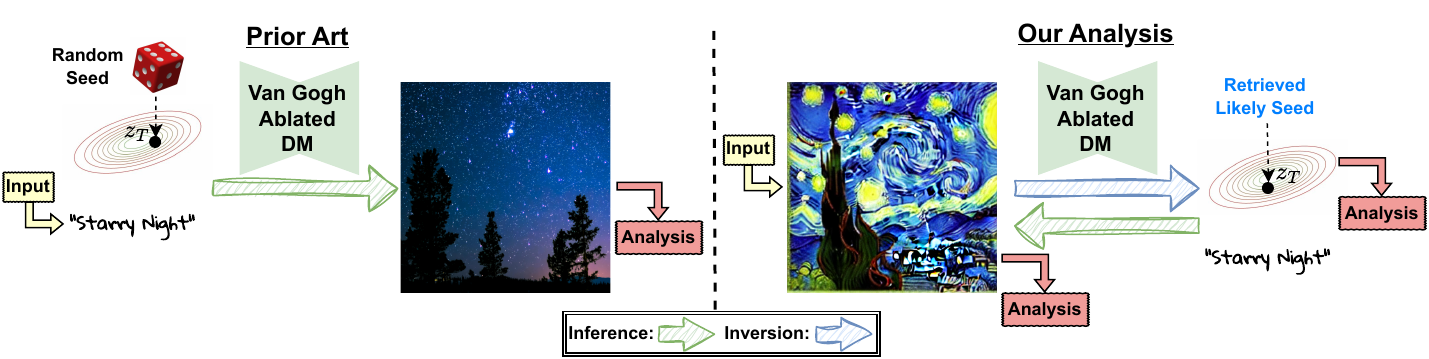}
    }

        \caption{
\textbf{Evaluation of concept erasure models: Prior Art vs. Our Analysis.} Prior art analyzes the image generated by an ablated model using the text (or textual embeddings) and a random seed. Instead, we assume that both text and ablated image are given and analyze the likelihood of the corresponding seed, in the latent space of the model, as well as the quality of the generated image. We find that ablated models contain seeds with high likelihood that can be used to generate high quality ablated images.}
    \label{fig:teaser}
    \vspace{-9pt}
\end{figure*}

%% file: sections/02_related_work.tex
\section{Background}
\label{sec:background}

\subsection{Diffusion models concept erasure}
In recent years, diffusion models~\cite{ddpm,diffusion_sohl_dickstein} have demonstrated significant advancements in image generation. Further improvements~\cite{ho2022cfg,rombach2022ldm,dhariwal2021diffusion,rombach2022ldm} allowed high fidelity text-to-image generation.
These models are trained on large-scale datasets containing images from a wide variety of categories.

Post-training, several issues regarding the models' generation may arise. For example, some images in the dataset contain not-safe-for-work (NSFW) content, copyrighted images, or private content, which the models learn to generate.

A possible way to remove the effect of certain training data on the model is to retrain the model from scratch excluding that data. 
However, as these issues can recur multiple times on large-scale models, often retraining is infeasible.
Moreover, 
problematic images might be generated even if they are not part of the training set~\cite{iwf_csam}.

These concerns raise the need for techniques that can edit a diffusion model to change its outputs \wrt given data.
Given a pre-trained model, the process of removing the effect of training data from it is referred to as \textit{machine unlearning}~\cite{machine_unlearning}. 
This has been explored vastly for discriminative tasks~\cite{certified_data_removal,when_unlearning_jeopardizes,eternal_spotless,tanno2022repairing,wu2022puma}, as the effect of data samples on models prediction is more direct in this case.

As generative models have gained vast popularity recently, many unlearning issues and concerns arise for these models too, \eg, privacy regulations~\cite{GDPR} and generation of NSFW content~\cite{schramowski2023sld}.
For image generation tasks, earlier architectures have been examined~\cite{Malnick_2024_WACV,kong2023data_redaction}, with more recent studies focusing on diffusion models~\cite{kumari2023ablating_concepts,wu2024erasediff,gandikota2023esd,zhang2024fmn,fan2023salun,wu2024scissorhands,lyu2024spm,gandikota2024uce,schramowski2023sld,heng2024selective}. Schramowski \etal~\cite{schramowski2023sld} propose modifying a model's inference behavior to limit generation of certain data. Other methods~\cite{kumari2023ablating_concepts,wu2024erasediff,gandikota2023esd,zhang2024fmn,fan2023salun,wu2024scissorhands,lyu2024spm,gandikota2024uce,lu2024mace} suggest to finetune the model to reach this goal, or focus on changing the textual embeddings~\cite{zhang2024advunlearn,pham2024robust}.

Recently, there have been works that question and quantify the erasing concepts and abilities that these methods possess.
Zhang \etal~\cite{zhang2025unlearndiffatk} show an attack method against these models, using adversarial prompts that lead to the generation of an erased concept. Unlike this work, we do not propose an attack on concept erasing methods, nor do we aim to find specific prompts that generate concept images. Instead, we invert a concept image to find a suitable latent, showing that the image still lies in the plausible region of the distribution, even after the erasure process.

The closest work to ours is
Pham \etal~\cite{pham2024circumventing}, examining different concept erasing methods by using textual inversion~\cite{gal2022image} to find a suitable textual embedding for generating given erased concept images.
As opposed to their work, we focus on retrieving $z_T$ latents that can produce the concept image and {\em analyze their likelihood}.

\subsection{Latent Diffusion Models}
\label{subsec:ldm}
Diffusion models~\cite{ddpm,diffusion_sohl_dickstein} are generative models that learn to map Gaussian noise $x_T$ to an image $x_0$ in a gradual denoising process over multiple timesteps $t \in [0, T]$. This is done by training a learnable neural network $\epsilon_{\theta}(\cdot, \cdot)$ that learns to reverse a known forward Markov chain with Gaussian noise transitions with predefined parameters $\alpha_t$. This means that given $\epsilon \sim \mathcal{N}(0, I)$, $x_t$ can be parameterized as:

\begin{equation}
\label{eq:diffusion_from_x_0}
x_t = \sqrt{\alpha_t}x_0 + \sqrt{1 - \alpha_t}\epsilon.    
\end{equation}
And for generation, $x_{t-1}$ can be expressed using the network's output:
\begin{equation}
    x_{t-1} = \sqrt{\frac{\alpha_{t-1}}{\alpha_t}} \cdot x_t - \gamma_t(\alpha_t, \alpha_{t-1}) \cdot \epsilon_\theta(x_t, t),
\end{equation}

where $\gamma_t(\alpha_t, \alpha_{t-1})$ is a noise variance parameter.
During training, the model learns to predict the added noise $\epsilon$. This means the loss is:
\begin{equation}
    \label{eq:dm_loss}
    \mathcal{L}_{\mathrm{DM}}:= \displaystyle {\mathop{\mathbb{E}}_{x, \epsilon, t} \left[ \Pnorm{\epsilon - \epsilon_\theta(x_t, t)}{2}^2 \right]}.
\end{equation}
Further advancements~\cite{ho2022classifier_free_guidance,dhariwal2021diffusion} have allowed conditioning the generation on textual prompts, allowing the model to receive an additional text $c$ as an input, \ie $\epsilon_{\theta}(x_t, t, c)$. For faster computing in space with lower complexity than the image space, 
Rombach \etal~\cite{rombach2022ldm} showed that using a VAE~\cite{variational_auto_encoder} encoder and decoder, denoted as $\textrm{Enc}(\cdot)$ and $\textrm{Dec}(\cdot)$, respectively, the memory efficiency of the diffusion process can improve. Instead of training the diffusion process on the high dimensional image $x_0$, we encode the image to a lower latent space, \ie, $\textrm{Enc}(x_0) = z_0$, with $\textrm{dim}(z_0) << \textrm{dim}(x_0)$. Then, the diffusion process is done in this lower space. This model is denoted as a \textit{Latent Diffusion Model} (LDM).
The loss term for training LDMs is thus:
\begin{equation}
    \label{eq:ldm_loss}
    \mathcal{L}_{\textrm{LDM}} := \displaystyle {\mathop{\mathbb{E}}_{\textrm{Enc}(x),\epsilon, t} \left[ \Pnorm{\epsilon - \epsilon_\theta(z_t, t, c)}{2} \right]}.
\end{equation}
Generation is done by sampling a latent seed $z_T\sim \mathcal{N}(0, I)$ and using the denoising network $\epsilon_\theta(\cdot, \cdot, \cdot)$ to iteratively compute $z_0$. Then, the output image is produced by the decoder, \ie,  $\hat{\mathcal{I}} = \textrm{Dec}(z_0)$. In our work, we specifically focus on LDMs, as we use the decoder in some of our analyses. We refer to the process of generating an image from a latent $z_T$ as {\em diffusion inference}.

\input{figures/12_nll_histogram/figure}
\input{figures/22_emd_ratio_explain/figure}
\subsection{Diffusion model inversion}

For diffusion models, inversion is the procedure of finding the latent seed that can be used to generate a given image. As the generation nature of diffusion models is iterative, simple optimizations are too computationally heavy to perform on SOTA models. 
As DDIM sampling~\cite{song2020ddim} can be used deterministically, DDIM inversion~\cite{dhariwal2021diffusion} was proposed as a simple way to invert. 
Subsequent work~\cite{mokady2023null_text_inversion} has shown that this simple method can produce inferior results and proposed a method to invert an image by optimizing for a better null text token embedding. Additional studies~\cite{miyake2023negative,samuel2023regularized,wallace2023edict} have also proposed alternative inversion methods, showing results with very low reconstruction errors. Garibi \etal~\cite{garibi2024renoise} proposed a method that inverts an image by using iterative steps of refinement between the diffusion steps, termed Renoise.

%% file: figures/12_nll_histogram/figure.tex
\begin{figure}[t]
    \centering
    \resizebox{0.48 \textwidth}{!}{
    \includegraphics[]{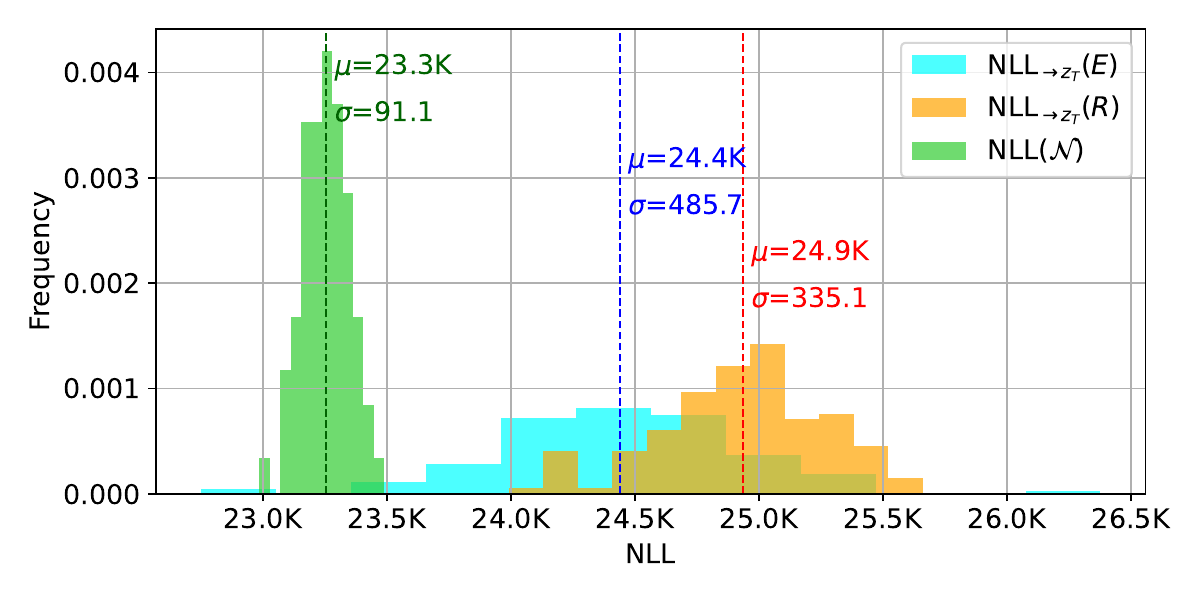}
    }
        \caption{\textbf{NLL histogram:} For a model that erased the concept Nudity (EraseDiff~\cite{wu2024erasediff}), the likelihood distribution fits different Gaussians (\textcolor{cyan_real}{$\nllzt(E)$}, \textcolor{orange}{$\nllzt(R)$}), that are different from the sampling distribution of the LDM which is standard normal distribution (\textcolor{LimeGreen}{$\text{NLL}(\mathcal{N})$}).} 
        
    \label{fig:nll_histogram}
\end{figure}

%% file: figures/22_emd_ratio_explain/figure.tex
\begin{figure}
    \centering
    
    \resizebox{1\linewidth}{!}{{\includegraphics[]{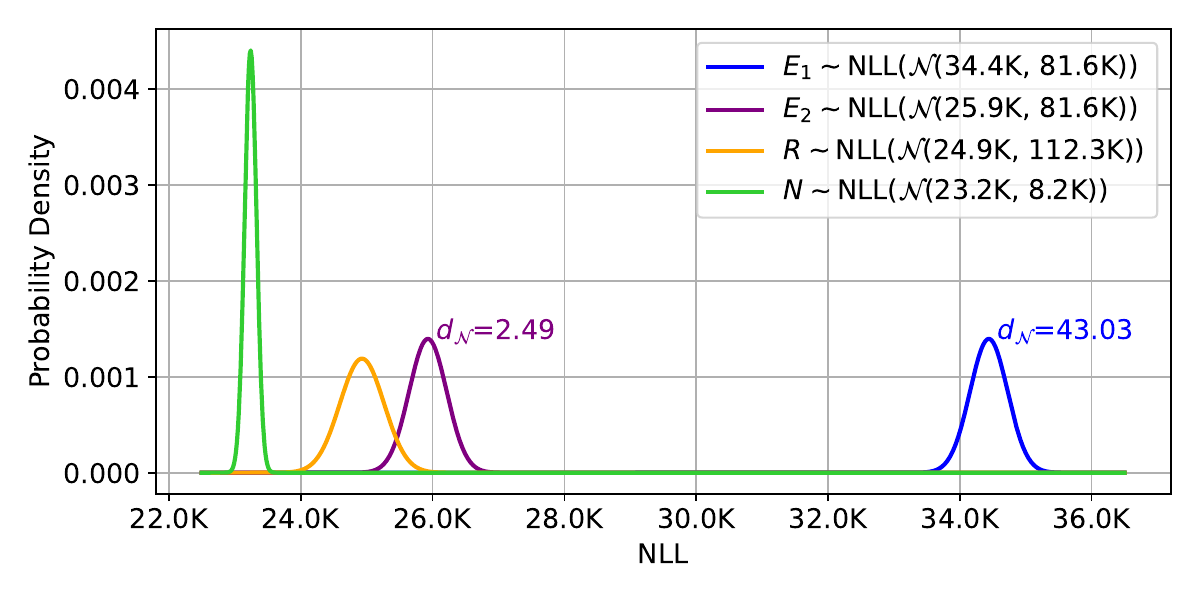}}}

    \caption{
    \textbf{Visualizing our distance measure:}
    Our {\em relative distance} measure is the ratio of $EMD(E,\mathcal{N}))$ to $EMD(R,\mathcal{N}))$, where $E$ is the {\em erased} set, \textcolor{orange}{$R$} is the {\em reference} set, \textcolor{LimeGreen}{$\mathcal{N}$} is the normal distribution, and EMD is Earth Movers Distance. As can be seen, the erased model \textcolor{blue}{$E_1$} is much farther than \textcolor{purple_real}{$E_2$}, suggesting that the model that forgot \textcolor{blue}{$E_1$} did a much better job.}

    \label{fig:emd_explain}
\end{figure}

%% file: sections/03_analysis.tex
\input{figures/11_concept_flow/figure}
\input{figures/17_psnrs_all_in_one/figure}

\section{Analysis}
\label{sec:analysis}
\paragraph{Basic setup.} To evaluate a model that erased a given concept $c$, we require the following:
\begin{enumerate}
    \item White box access to the LDM model that erased $c$, denoted $\epsilon_\theta^c$.
    \item An \emph{erased set} $E$ of (image, caption) pairs with images containing the concept $c$.
    \item A \emph{reference set} $R$ of (image, caption) pairs with images that do not contain the concept $c$.
\end{enumerate}

Our goal is to analyze and quantify the erasing effect of $\epsilon_\theta^c$ \wrt the concept $c$. This is done by retrieving a latent vector that, along with $\epsilon_\theta^c$, can be used to generate images containing $c$. We require that the images have a low reconstruction error, \ie, high PSNR, and analyze the likelihood of the latent vector.

\subsection{How do we measure memory?}
\label{subsec:measure_likelihood}

Given a latent seed $z_T$, we are interested in examining its likelihood. As the distribution that was used to train the model is Gaussian (see \cref{subsec:ldm}), \ie, $\mathcal{N}(0, I)$ or $\mathcal{N}$ in short, computing the likelihood is straightforward via the closed-form probability density function (PDF). We report our results in Negative Log Likelihood (NLL) units, denoting the NLL of a given normal distribution with parameters $(\mu, \sigma^2I)$ as $\text{NLL}(\mathcal{N}(\mu, \sigma^2I))$. 
We follow the Central Limit Theorem to approximate $\text{NLL}(\mathcal{N}(\mu, \sigma^2\cdot I))$ as 1D Gaussian (see Supplementary for a detailed analysis).

For a set of images, we invert them to latent seeds $z_T$ (see~\cref{subsec:memory_of_concept,subsec:memory_of_image}) and compute their NLL. We denote this function as $\nllzt(\cdot)$. We perform this on the erased and reference sets, denoting these distributions as $\nllzt(E)$ and  $\nllzt(R)$, respectively. These distributions can be seen in~\cref{fig:nll_histogram} as histograms, showing the separation of the likelihood of latents from different populations.

We found it difficult to have a clear understanding based on the NLL values alone (as in~\cref{fig:nll_histogram}). Therefore, we opt for a unit-less number that conveys information in relative terms.
From a likelihood perspective, a model that erases a concept should ideally map it to a low-likelihood region in $z_T$ space while preserving non-erased concepts. Therefore we measure the distance of images in the erased set $E$ to the normal distribution in terms of the distance of the reference set $R$. Specifically, we use the ratio of the Earth Mover’s Distance (EMD)~\cite{rubner2000earth_movers_distance} to obtain this measure, denoted {\em Relative Distance}:
\begin{equation}
\label{eq:relative_similarity}
    d_{\mathcal{N}} (E, R) := \frac{\text{EMD}\left( \nllzt (E) ,\text{NLL}(\mathcal{N})\right)}{\text{EMD}\left(\nllzt(R),\text{NLL}(\mathcal{N})\right)}.
\end{equation}

The measure $d_{\mathcal{N}}(\cdot, \cdot)$  should approach $1$ if the two distances are roughly the same (a value smaller than $1$ suggests the erased set is closer to the Normal distribution than the reference set, which indicates that something is wrong with the model). A high $d_{\mathcal{N}}(\cdot, \cdot)$ means the reference set is far more likely than the erased set, which is what we hope for.

\cref{fig:emd_explain} illustrates the two different outcomes of $d_{\mathcal{N}}(\cdot, \cdot)$ for the distributions of different sets, $E_1$, $E_2$, $R$, along with $\mathcal{N}$ for a standard normal distribution ($\text{NLL}(\mathcal{N})$). For simplicity, we use different normal distributions for $\nllzt(E_1), \nllzt(E_2), \nllzt(R)$.
We see that $E_1$ is far from both $R$ and $\mathcal{N}$, with very high relative distance of $d_{\mathcal{N}}(R, E_1) = 43.02$. In contrast, $E_2$ is closer to both these distributions, overlapping $R$ with a low relative distance of $d_{\mathcal{N}}(R, E_2) = 2.49$. 
The high score of $d_{\mathcal{N}}(R, E_1)$ suggests that the images in $E_1$ are much less likely for generation while preserving a small distance between the reference set and the standard normal distribution.

\paragraph{Experimental setup.} We now show that the information of ablated concepts persists in erased models.
To achieve this, we inquire into six different concepts: Nudity, Van Gogh, Church, Garbage Truck, Parachute and Tench.
We choose a set of nine different erasure methods, each accompanied by the models used to erase the concepts above, as reported in their original publications:
ESD~\cite{gandikota2023esd}, FMN~\cite{zhang2024forget},
SPM~\cite{lyu2024spm} and
AdvUnlearn~\cite{zhang2024defensive} on all concepts.
Salun~\cite{fan2023salun}, Scissorhands~\cite{wu2024scissorhands},
and 
EraseDiff~\cite{wu2024erasediff} on all concepts apart from Van Gogh.  UCE~\cite{gandikota2024uce} on Nudity and Van Gogh.
AC~\cite{kumari2023ablating} only on Van Gogh.
All models ablate the base Stable-Diffusion v1.4~\cite{rombach2022stable_diffusion}.
The latent space dimension for this model is $4\times 64\times 64$.
 We follow the evaluation protocol of Zhang \etal~\cite{zhang2025unlearndiffatk} and collect the datasets in the same manner. For NSFW content, we use the I2P dataset~\cite{schramowski2023sld}.

For the reference set $R$ in our analysis, we use images from the COCO~\cite{lin2014microsoft_mscoco} dataset.
We use Renoise~\cite{garibi2024renoise} as our primary inversion method, with the captions attached to the images as guidance, for $50$ inversion steps and $5$ renoising steps (see Supplementary for more information on parameter choice).

We consider two types of analyses. The first is inquiring about the memory of an ablated concept, by aiming to retrieve a single latent to every given query image. The second, exploring many memories of an ablated image, aiming to find multiple seeds that correspond to the same query image. Following previous works in the field, we also report additional metrics regarding the generated images in the Supplementary, including CLIP~\cite{radford2021clip} score for prompt-image alignment and concept detection scores.

\subsection{Memory of an ablated concept}
\label{subsec:memory_of_concept}
Equipped with our Relative Distance measure, we show that on a dataset level, concept erasure models can generate the erased concepts with high PSNR and high likelihood. Namely, for every image in the dataset, we can find a likely latent that recovers that image.

Given the erased set with images that depict the concept $c$ (\eg, various images of churches), namely $E = \{(\mathcal{I}_i, p_i)\}_{i=1}^n$, we use it to analyze a model that was finetuned to erase this concept, $\epsilon_\theta^c$ (see~\cref{subsec:measure_likelihood}). 

For every query image $(\mathcal{I}_q, p_q)\in E$, we perform diffusion inversion to retrieve a latent seed $z_T^q$. This latent is later used for diffusion inference, leading to a reconstructed image $\hat{\mathcal{I}_q}$.
\cref{fig:concept_analysis} illustrates this procedure.
$z^q_T$ and $\hat{\mathcal{I}_q}$ are both used for our analysis.

\paragraph{Results.}
The results, summarized in~\cref{fig:concept-level-results}, demonstrate how all current concept erasing methods can generate images containing the erased concept. This is suggested by the high PSNR values for reconstruction among all the different methods.
Concepts with finer texture details such as Van Gogh and Church, tend to have lower PSNR, while smoother concepts such as Nudity and Parachute, achieve higher reconstruction. For example, Van Gogh images contain many brushstrokes, while Church images feature many bricks with distinct contours. In contrast, Nudity images have smooth surfaces, while Parachute images have large areas of background with similar color values.

The right panel of~\cref{fig:concept-level-results} shows the relative distance ($d_\mathcal{N}(\cdot, \cdot)$) of different concepts and concept erasing methods.
First, we see that the highest distance, meaning the best distance in terms of erasing, is achieved by ESD~\cite{gandikota2023esd} on the concept $c = $ Parachute, with a distance of $d_{\mathcal{N}}(E_c, R_c) = 2.49$.
This score indicates that there exists a non-negligible overlap between the distribution of the erased concept and the reference dataset. Please refer to~\cref{fig:emd_explain} for a visualization of a $2.49$ distance. The Supplementary contains additional examples and results, showing how similar the reconstructed images $\hat{\mathcal{I}}_q$ are to the concept images $\mathcal{I}_q$.

\input{figures/14_image_flow/figure}
\input{figures/19.1_image_level_recalling_collage_2/figure}

Moreover, compared to the Vanilla model which did not erase the given concept, many methods achieve a similar relative distance $d_\mathcal{N}(\cdot, \cdot)$. This suggests that while these methods exemplify that it is harder to generate images of the erased concept using text prompts that describe it, the generation of such images in the latent space is still plausible. Observe that in some scenarios the distance is lower than 1, which seems to suggest that these models erased a concept via a text proxy but did not actually forget it. %

\input{figures/18_image_forget_psnr_and_nll_score_bar_plots/figure}
\input{figures/23_average_cosine_distance/figure}

\subsection{The many memories of an ablated image }
\label{subsec:memory_of_image}

In the previous subsection, we analyzed the case of erasing a concept using multiple images, by finding a feasible latent $z_T$ that can be used to generate an image that depicts the concept.
However, this raises the question: for a given image $\mathcal{I}_q$, is there more than one distinct latent seed $z_T^q$ that can generate an image that resembles $\mathcal{I}_q$?
Specifically, we are interested in whether we can find several $z_T$ latents with a sufficiently large cosine distance between them, that all can be used to generate the query image $\mathcal{I}_q$. These latent vectors should satisfy two main requirements: they should be likely (in terms of the model’s probability distribution) and should be well-separated from each other. Specifically, we are interested in distant memories of the query image, and not adjacent ones.
\cref{fig:image_analysis} illustrates our approach.

\paragraph{Sequential Inversion Block}
We seek distinct ``\textit{memories}" of the same query image. To do that, we start with random support images. (A detailed analysis of alternative initialization choices beyond images can be found in the Supplementary.) For each support image, we invert the VAE decoder $\textrm{Dec}(\cdot)$, to obtain an initial latent vector $z_0$ which is then used for an optimization process \wrt the query image.
Similarly to~\cref{subsec:memory_of_concept}, the reconstructed query images are fed to the diffusion inversion to produce the desired $z_T$ seeds, which we call distinct ``memories". 

Formally, we introduce the \textit{Sequential Inversion Block}, which maps support images $\mathcal{I}_{s_i}$ from the image space to the latent space. 
This is done using the following sequential inversion steps:

\begin{enumerate}[itemsep=0pt, parsep=0pt]
    \item \textbf{Initial Decoder Inversion:} 
     
    Find an initial latent, $z_0^{(s_i)}$, in the decoder's latent space, that will serve as a starting point for the next step in the block.
    Specifically, invert the VAE decoder $\textrm{Dec}(\cdot)$, starting from $\textrm{Enc}(\mathcal{I}_{s_i})$, and optimize the following:
    \begin{equation}
    \label{eq:vae_decoder_inversion_init}
        z_0^{(s_i)} = \argmin_{z} 
        \left[\text{Dist}\left(\textrm{Dec}\left(z\right), \mathcal{I}_{s_i}\right)\right] ,
    \end{equation}
    where $\text{Dist}(\cdot, \cdot)$ is the euclidean distance.
    \item \textbf{Decoder Inversion Towards the Query Image:}
    Next, starting from an initial latent $z_0^{(s_i)}$ (which corresponds to the support image $\mathcal{I}_{s_i}$), we optimize to find a latent that reconstructs the query image $\mathcal{I}_q$:
    \begin{equation}
    \label{eq:vae_decoder_inversion_towards}
        z_0^{(s_i \rightarrow q)} = \argmin_{z} 
        \left[\text{Dist}\left(\textrm{Dec}\left(z\right), \mathcal{I}_q\right)\right] .
    \end{equation}
    \item \textbf{Latent Diffusion Inversion:} We use $z_0^{(s_i \rightarrow q)}$ as a starting point for a diffusion model inversion process, resulting in a latent seed $z_T^{(s_i \rightarrow q)}$ which is the output of the \textit{Sequential Inversion Block}. 
\end{enumerate}

The retrieved $z_T^{(s_i \rightarrow q)}$ latent will also be used to generate an image $\hat{\mathcal{I}}_q$ that resembles ${\mathcal{I}}_q$ for reconstruction quality analysis. In addition, we analyze the likelihood of $z_T^{(s_i \rightarrow q)}$ and measure the average pairwise cosine distances between the generated $z_T^{(s_i \rightarrow q)}$, for all support images $\mathcal{I}_{s_i}$, to ensure we found distant seed vectors in latent space that represent the same $\mathcal{I}_q$ image.

\vspace{-7pt}
\paragraph{Experimental Setting: }
We extend the experimental setup in~\cref{subsec:measure_likelihood}. 
We randomly select support images from the COCO~\cite{lin2014microsoft_mscoco} dataset.
For each concept, we randomly choose five query images and validate that for every query image, there are at least ten distinct latents. Each of these latents is initialized using a different support image.
We set the number of VAE decoder optimization steps to 3,000.

\vspace{-7pt}
\paragraph{Results:}

A qualitative example of reconstructed images that were produced from this procedure is presented in ~\cref{fig:image_collage}, demonstrating how different latents can generate images of an erased concept.
\cref{fig:image-level-results} demonstrates the results, showing that for all methods and all concepts we were able to recover likely latent seed vectors, \ie, low relative similarity (\cref{eq:relative_similarity}) that lead to a high-quality reconstruction (high PSNR). 
However, compared to the concept level forgetting, (see~\cref{fig:concept-level-results}), we see a lower PSNR value and lower
$d_{\mathcal{N}}(\cdot, \cdot)$ values, when multiple diverse $z_T$ seeds are searched for. 
We conjecture that the search for multiple latents seeds is suboptimal to finding a single latent in an unconstrained setting

Additionally, we measure the distances between all latents within a concept. Specifically, we evaluate the cosine distance and euclidean distance (in the Supplementary) between all $z_T^{(s_i \rightarrow q_j)}$ latents associated with a given concept. This metric provides insight into the distribution of latents. These distances are presented in~\cref{fig:avg_cos_dist}. 

\vspace{-8pt}
\paragraph{Geometric interpretation of the retrieved memories:}
To better understand how the retrieved $z^{(s_i \rightarrow q)}_T$ seeds are distributed geometrically in space with respect to the original target $z_T^q$ seed, we compute the average euclidean distance between all $z^{(s_i \rightarrow q)}_T$ and $z^q_T$. Specifically, we refer to the illustration presented on the left part of \cref{fig:image_analysis}. We observe that all these distances are tightly spread around the mean distance. Concretely, the mean is $152.14$ and the standard deviation is $2.72$. This leads to a coefficient of variation of $2\%$. We extend the computation of the coefficient of variation to all query images in all of our experiments, and observe that the mean and standard deviation of the coefficients are $2\%$ and $1\%$, respectively. 
Together with the observation that there exists a substantial cosine distance between these seeds (\cref{fig:avg_cos_dist}),
we conclude that for each target image, our procedure produces memories that lie (with a high probability) on a sphere centered around a $z_T$ seed that corresponds to that image.
\input{figures/08_inverting_scrambled_image/figure}
Following this geometric insight, one could choose any number of $N_S$ support images, and retrieve $N_S$ seeds using SIB. This raises following questions for future {\em forgetting} work: Will re-mapping all these possible seeds into images that do not resemble $\mathcal{I}_q$, be {\em sufficient} for forgetting? Is it {\em necessary} for forgetting?

%% file: figures/11_concept_flow/figure.tex
\begin{figure*}[h!]
    \centering
    \resizebox{1 \textwidth}{!}{
    \includegraphics[width=0.7\linewidth]{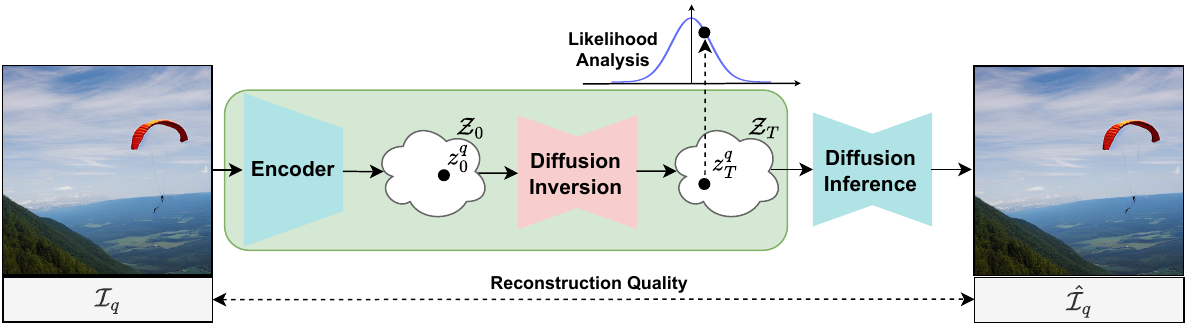}
    }
        \caption{
\textbf{Memory of an ablated image:}
 Given an ablated query image $\mathcal{I}_q$, our goal is to find a likely latent $z_T$ that can accurately reconstruct the image when processed through an ablated diffusion model. We start by encoding $\mathcal{I}_q$ into a latent $z_0$ with the encoder, then apply diffusion inversion to obtain a seed latent vector $z_T$. This seed is fed into the LDM to generate the image $\mathcal{\hat{I}}_q$. Finally, we evaluate the likelihood of $z_T$ and the quality of the reconstructed image $\mathcal{\hat{I}}_q$ compared to $\mathcal{I}_q$.}
    \label{fig:concept_analysis}
\end{figure*}

%% file: figures/17_psnrs_all_in_one/figure.tex
\begin{figure*}[h!]
    \centering
            \setlength{\tabcolsep}{0.8pt} %
        \renewcommand{\arraystretch}{0.6} %
    \begin{tabular}{c c}
    
    \resizebox{0.5\linewidth}{!}{{\includegraphics[]{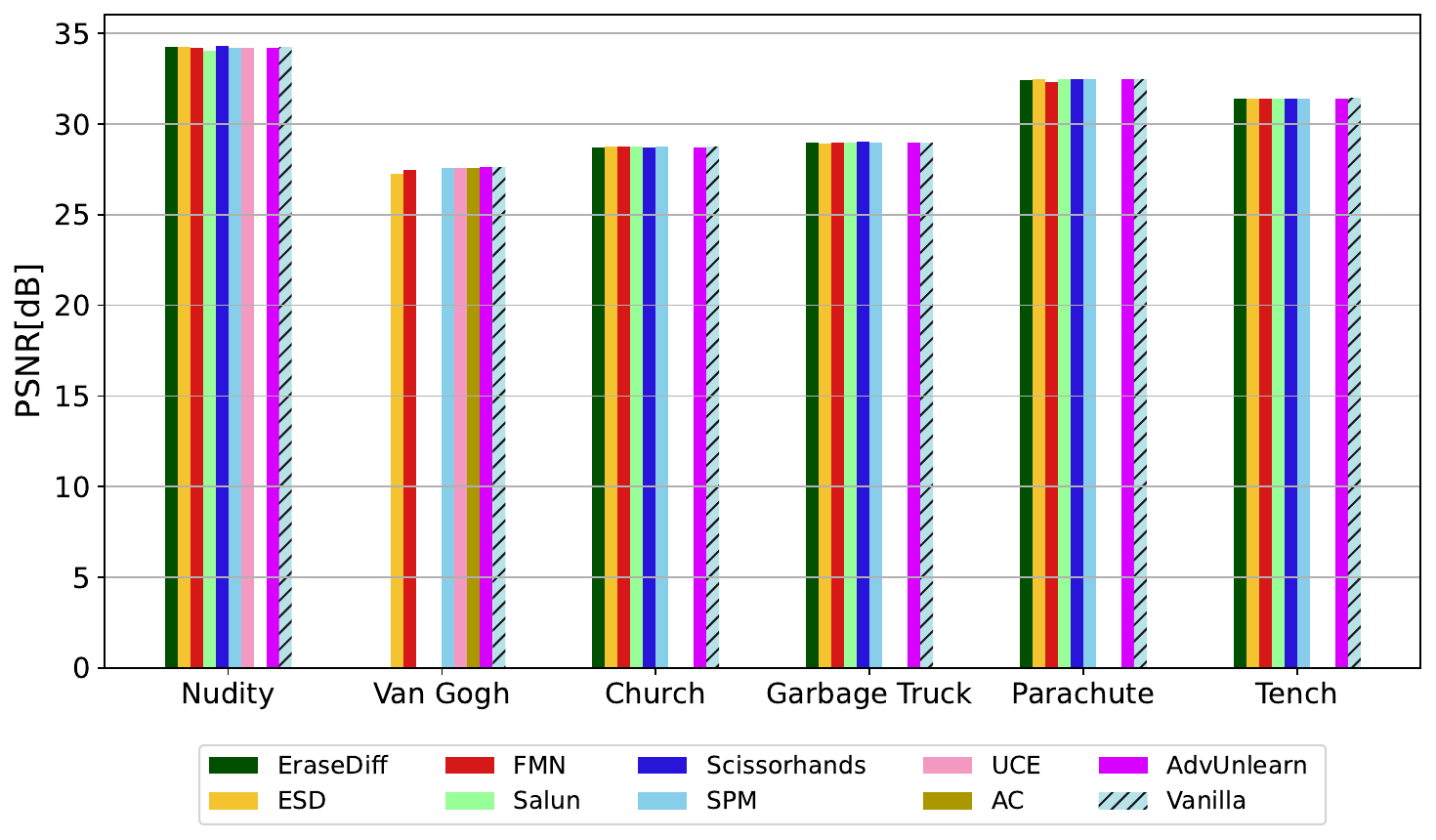}}} &
    \resizebox{0.5\linewidth}{!}{{\includegraphics[]{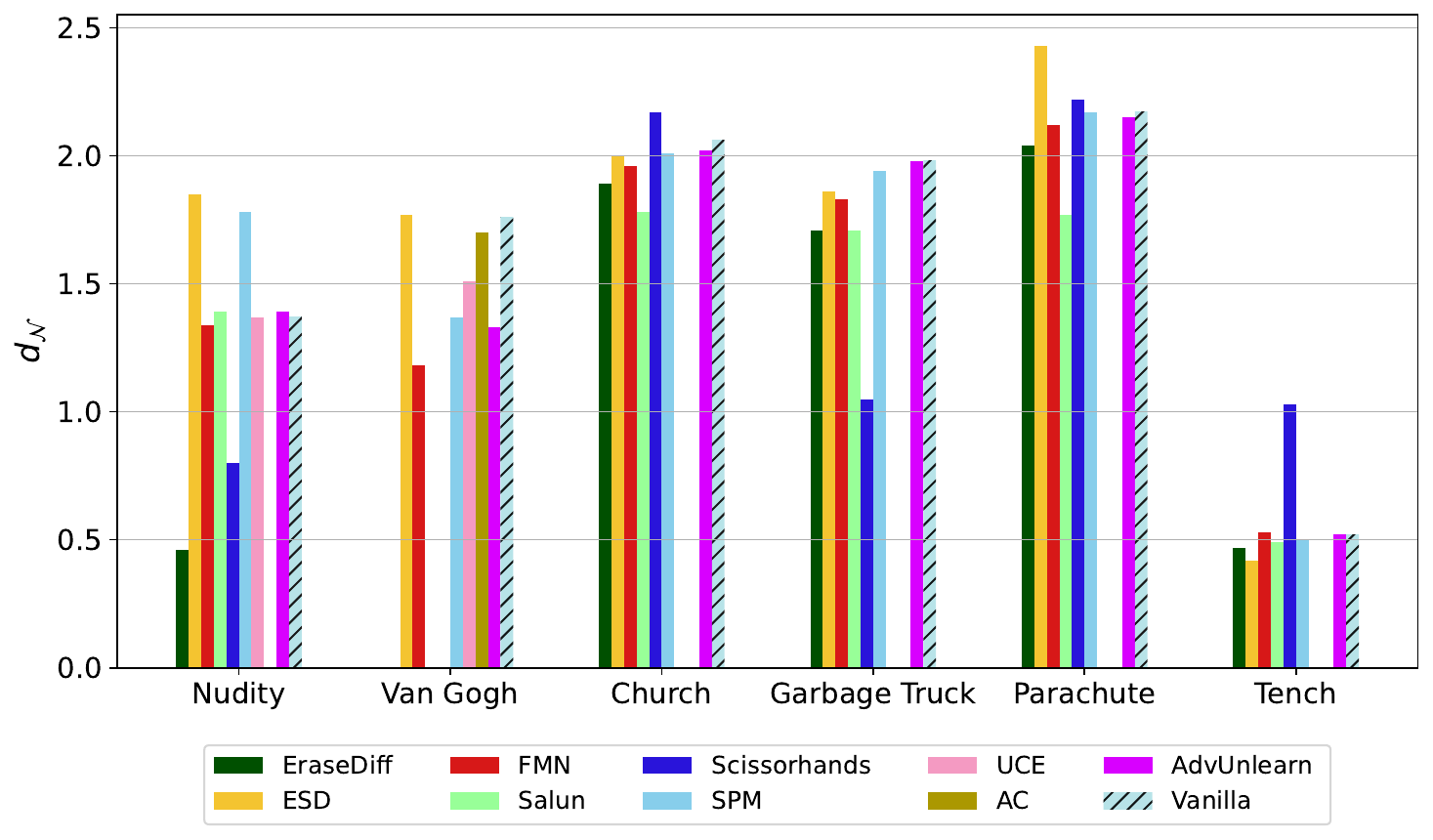}}}\\
    \scriptsize{(a) PSNR [dB] }& \scriptsize{(b) $d_{\mathcal{N}}(E, R)$ } 
    \end{tabular}
    
    \caption{
    \textbf{A concept erased model remembers:}
    We report the mean reconstruction PSNR (a) and our proposed relative distance (b) for six concept datasets $\{$Nudity, Van Gogh, Church, Garbage Truck, Parachute, Tench$\}$ across nine different concept ablation methods $\{$EraseDiff~\cite{wu2024erasediff}, ESD~\cite{gandikota2023esd}, FMN~\cite{zhang2024forget}, Salun~\cite{fan2023salun}, Scissorhands~\cite{wu2024scissorhands}, SPM~\cite{lyu2024spm}, UCE~\cite{gandikota2024uce}, AC~\cite{kumari2023ablating}, AdvUnlearn~\cite{zhang2024defensive}$\}$, along with one ``Vanilla" SD 1.4~\cite{Rombach_2022_CVPR} model. These results validate that, at the dataset level, there exists at least one latent per image that can reconstruct the image with high quality (PSNR $\geq 25$ dB) from a reasonable likelihood using the concept erased model.}
    \label{fig:concept-level-results}
\end{figure*}

%% file: figures/14_image_flow/figure.tex
\begin{figure*}[h]
    \centering
    \resizebox{1 \textwidth}{!}{
    \includegraphics[]{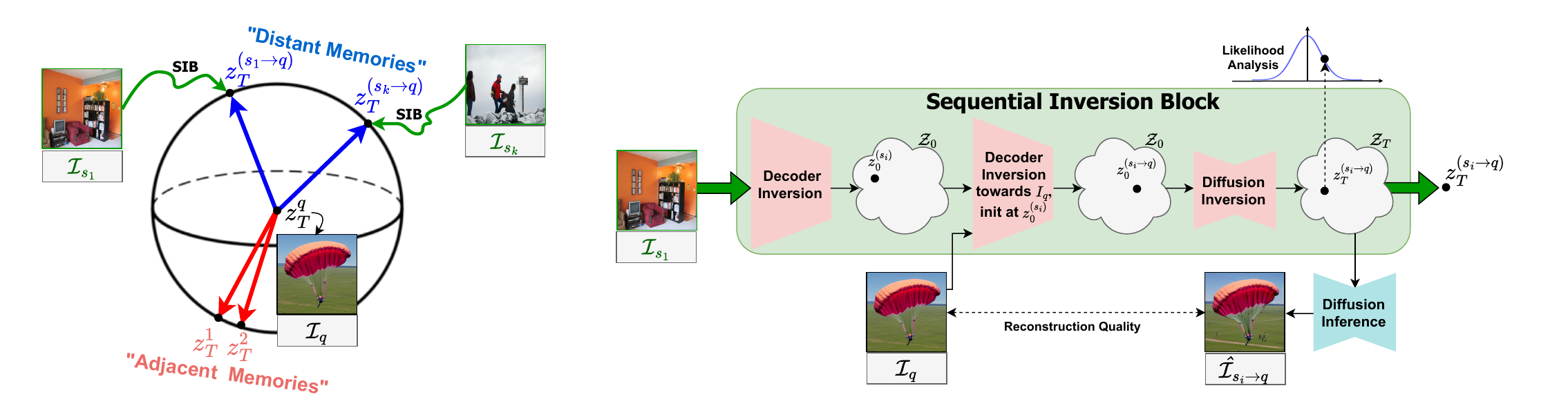}
    }

        \caption{
        \textbf{The many memories of an ablated image:}
        (Left) The latent seed $z_T^q$ of the query image $\mathcal{I}_q$ can be obtained from various support images: $\mathcal{I}_{s_1},...,\mathcal{I}_{s_k}$. For support image $\mathcal{I}_{s_i}$, we apply the sequential inversion block shown on the right to map it to the seed $z_T^{(s_i \rightarrow q)}$. We show in ~\cref{fig:concept-level-results} that seeds $\{z_T^{(s_i \rightarrow q)}\}_{i=1}^{k}$ are likely enough and can be used to generate the query image $\mathcal{I}_q$. (Right) Recovering the seed of a query image $\mathcal{I}_q$ when starting with support image $\mathcal{I}_{s_i}$.}

    \label{fig:image_analysis}
\end{figure*}

%% file: figures/19.1_image_level_recalling_collage_2/figure.tex
\begin{figure*}[h]
    \centering
    \setlength{\tabcolsep}{0.7pt} %
    \begin{tabular}{cccccp{5pt}ccccc}
    
        \multicolumn{5}{c}{{\small Support Images}} && \multicolumn{5}{c}{{\small Reconstructions}}
        \tabularnewline

        \adjustbox{valign=m}{\includegraphics[width=0.094\linewidth]{figures/19.1_image_level_recalling_collage_2/vangogh_support/0.png}} &
        \adjustbox{valign=m}{\includegraphics[width=0.094\linewidth]{figures/19.1_image_level_recalling_collage_2/vangogh_support/1.png}} &
        \adjustbox{valign=m}{\includegraphics[width=0.094\linewidth]{figures/19.1_image_level_recalling_collage_2/vangogh_support/2.png}} &
        \adjustbox{valign=m}{\includegraphics[width=0.094\linewidth]{figures/19.1_image_level_recalling_collage_2/vangogh_support/3.png}} &
        \adjustbox{valign=m}{\includegraphics[width=0.094\linewidth]{figures/19.1_image_level_recalling_collage_2/vangogh_support/4.png}} &

        &
        
        \adjustbox{valign=m}{\includegraphics[width=0.094\linewidth]{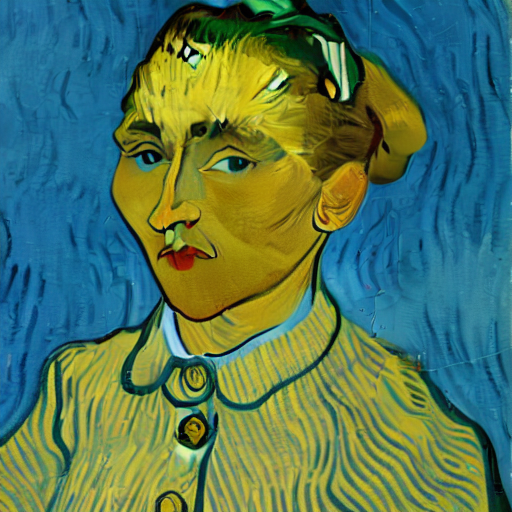}} &
        \adjustbox{valign=m}{\includegraphics[width=0.094\linewidth]{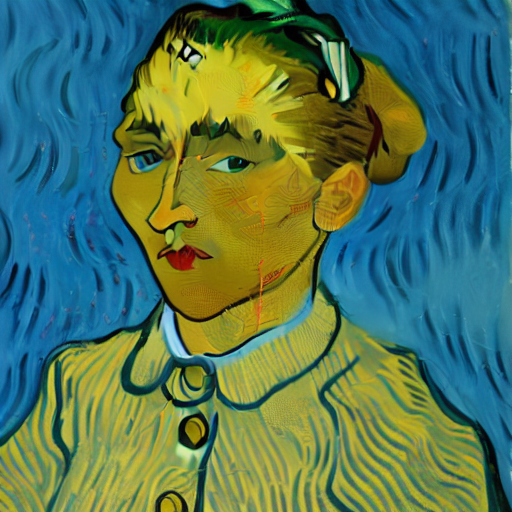}} &
        \adjustbox{valign=m}{\includegraphics[width=0.094\linewidth]{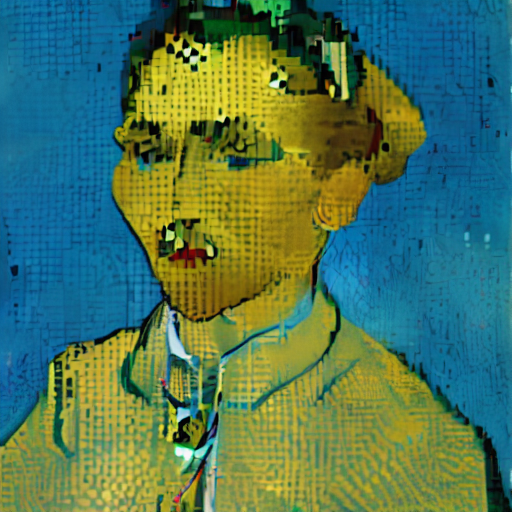}} &
        \adjustbox{valign=m}{\includegraphics[width=0.094\linewidth]{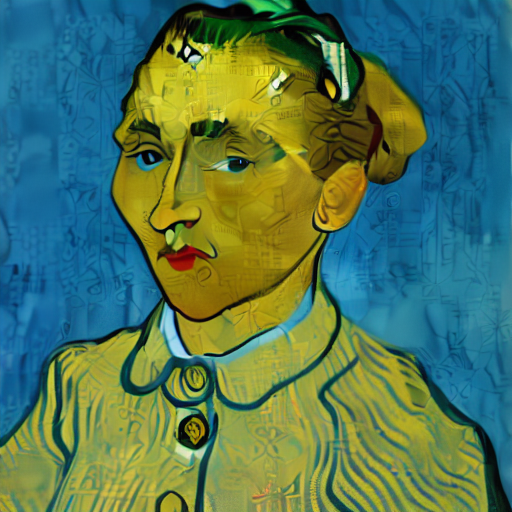}} &
        \adjustbox{valign=m}{\includegraphics[width=0.094\linewidth]{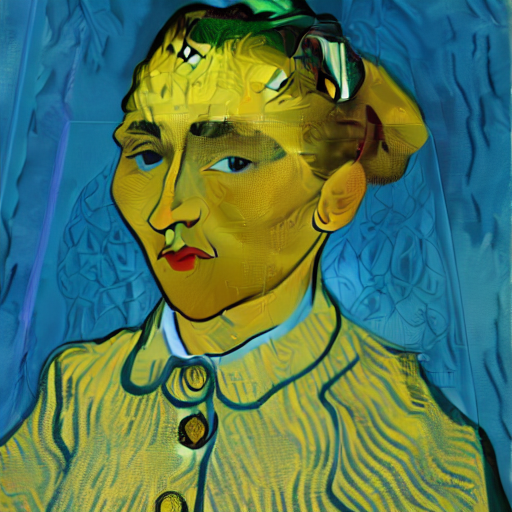}}

        \tabularnewline

        \adjustbox{valign=m}{\includegraphics[width=0.094\linewidth]{figures/19.1_image_level_recalling_collage_2/vangogh_support/5.png}} &
        \adjustbox{valign=m}{\includegraphics[width=0.094\linewidth]{figures/19.1_image_level_recalling_collage_2/vangogh_support/6.png}} &
        \adjustbox{valign=m}{\includegraphics[width=0.094\linewidth]{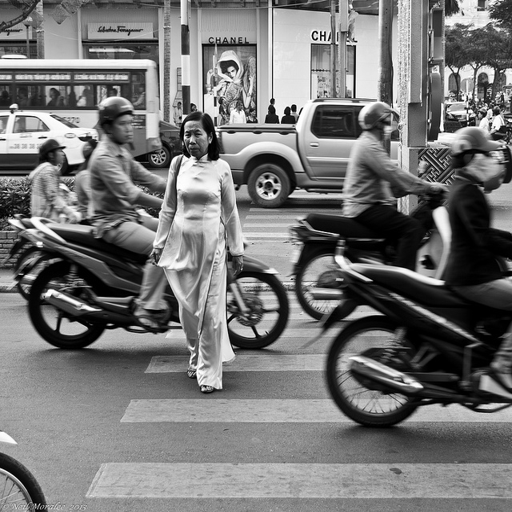}} &
        \adjustbox{valign=m}{\includegraphics[width=0.094\linewidth]{figures/19.1_image_level_recalling_collage_2/vangogh_support/8.png}} &
        \adjustbox{valign=m}{\includegraphics[width=0.094\linewidth]{figures/19.1_image_level_recalling_collage_2/vangogh_support/9.png}} &

        &
        
        \adjustbox{valign=m}{\includegraphics[width=0.094\linewidth]{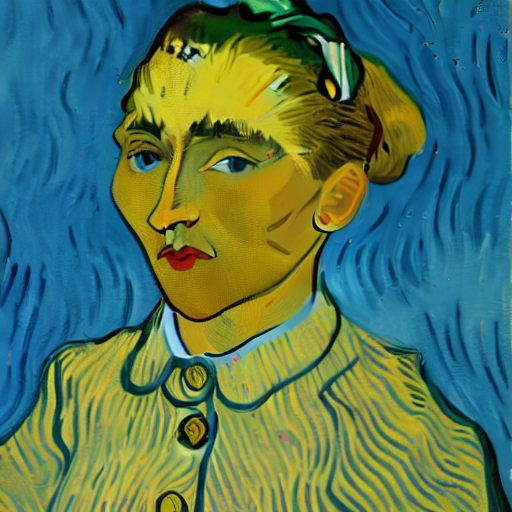}} &
        \adjustbox{valign=m}{\includegraphics[width=0.094\linewidth]{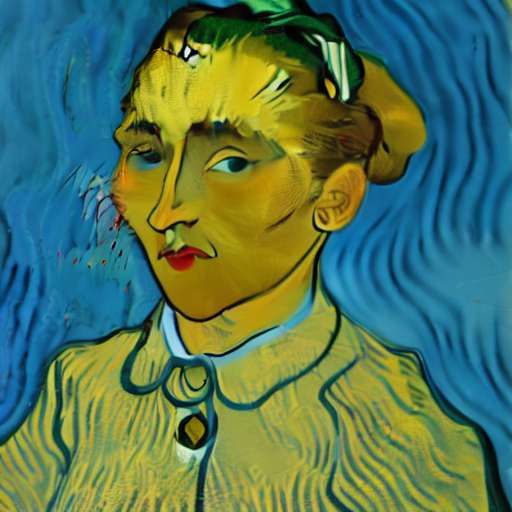}} &
        \adjustbox{valign=m}{\includegraphics[width=0.094\linewidth]{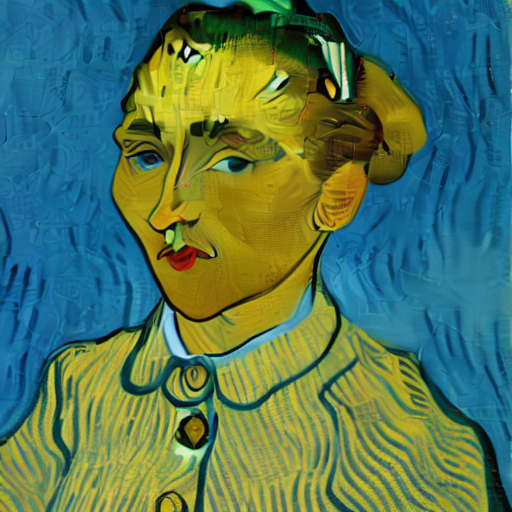}} &
        \adjustbox{valign=m}{\includegraphics[width=0.094\linewidth]{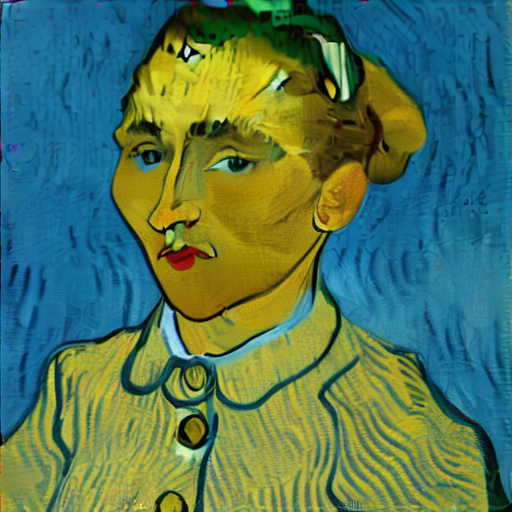}} &
        \adjustbox{valign=m}{\includegraphics[width=0.094\linewidth]{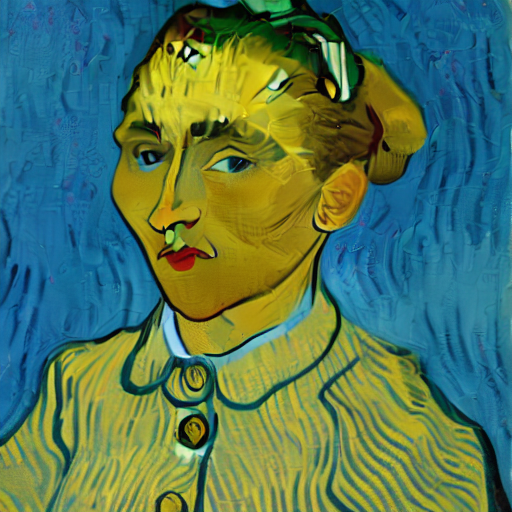}}
        
        \tabularnewline
        &&&&&&\multicolumn{5}{c}
        {\resizebox{0.48\linewidth}{!}{$\text{PSNR}_{\text{min}}=19.98[\text{dB}]$, $\text{PSNR}_{\text{avg}}=24.66[\text{dB}]$, $\text{PSNR}_{\text{max}} = 26.42[\text{dB}]$}}

        \tabularnewline

        \adjustbox{valign=m}{\includegraphics[width=0.094\linewidth]{figures/19.1_image_level_recalling_collage_2/tench_support/0.png}} &
        \adjustbox{valign=m}{\includegraphics[width=0.094\linewidth]{figures/19.1_image_level_recalling_collage_2/tench_support/1.png}} &
        \adjustbox{valign=m}{\includegraphics[width=0.094\linewidth]{figures/19.1_image_level_recalling_collage_2/tench_support/2.png}} &
        \adjustbox{valign=m}{\includegraphics[width=0.094\linewidth]{figures/19.1_image_level_recalling_collage_2/tench_support/3.png}} &
        \adjustbox{valign=m}{\includegraphics[width=0.094\linewidth]{figures/19.1_image_level_recalling_collage_2/tench_support/4.png}} &

        &
        
        \adjustbox{valign=m}{\includegraphics[width=0.094\linewidth]{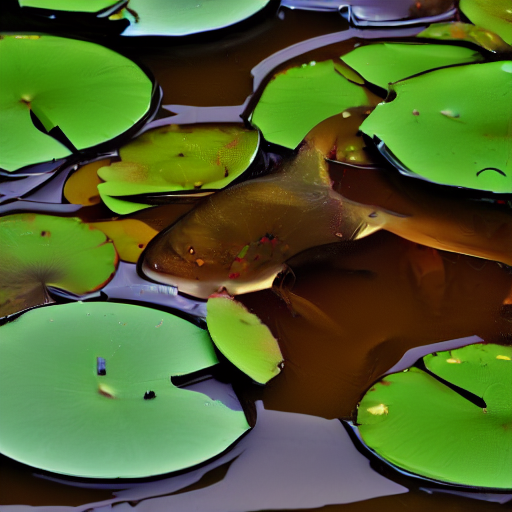}} &
        \adjustbox{valign=m}{\includegraphics[width=0.094\linewidth]{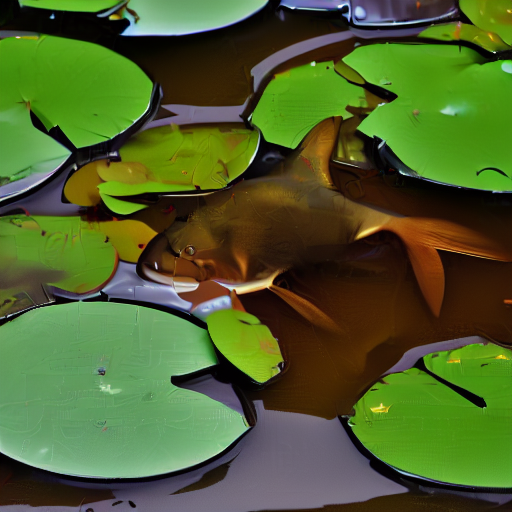}} &
        \adjustbox{valign=m}{\includegraphics[width=0.094\linewidth]{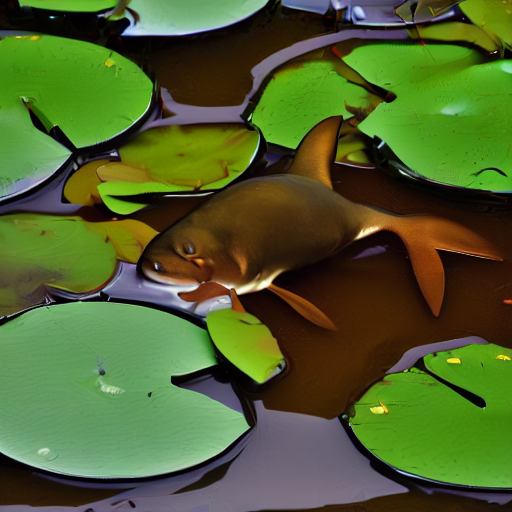}} &
        \adjustbox{valign=m}{\includegraphics[width=0.094\linewidth]{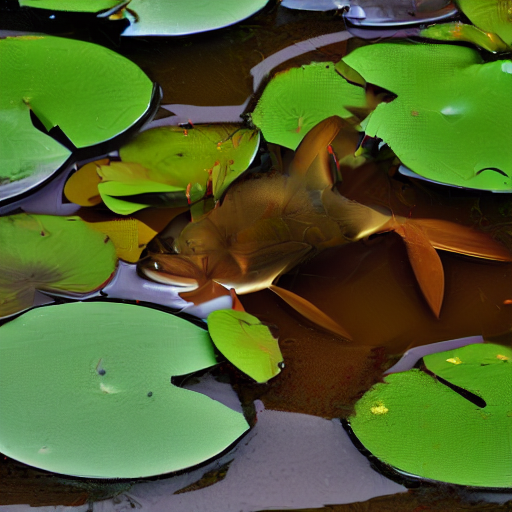}} &
        \adjustbox{valign=m}{\includegraphics[width=0.094\linewidth]{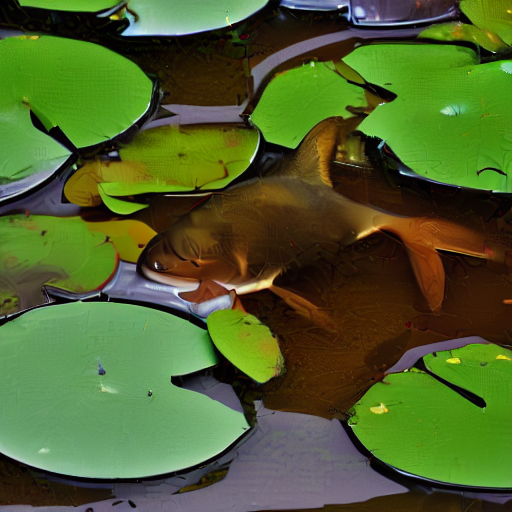}}
        
        \tabularnewline

        \adjustbox{valign=m}{\includegraphics[width=0.094\linewidth]{figures/19.1_image_level_recalling_collage_2/tench_support/5.png}} &
        \adjustbox{valign=m}{\includegraphics[width=0.094\linewidth]{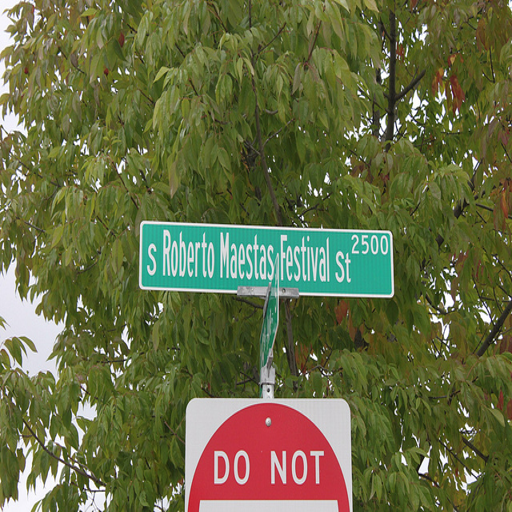}} &
        \adjustbox{valign=m}{\includegraphics[width=0.094\linewidth]{figures/19.1_image_level_recalling_collage_2/tench_support/7.png}} &
        \adjustbox{valign=m}{\includegraphics[width=0.094\linewidth]{figures/19.1_image_level_recalling_collage_2/tench_support/8.png}} &
        \adjustbox{valign=m}{\includegraphics[width=0.094\linewidth]{figures/19.1_image_level_recalling_collage_2/tench_support/9.png}} &

        &
        
        \adjustbox{valign=m}{\includegraphics[width=0.094\linewidth]{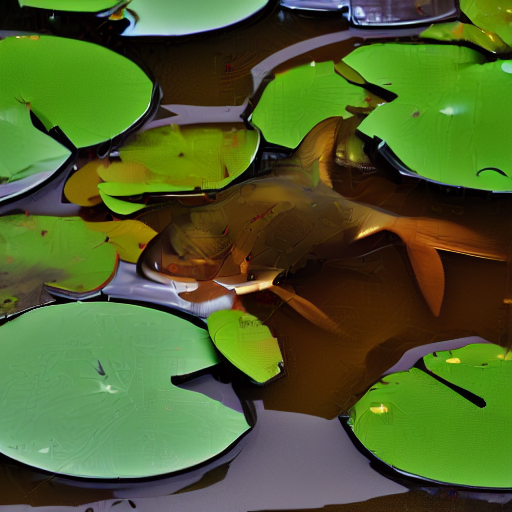}} &
        \adjustbox{valign=m}{\includegraphics[width=0.094\linewidth]{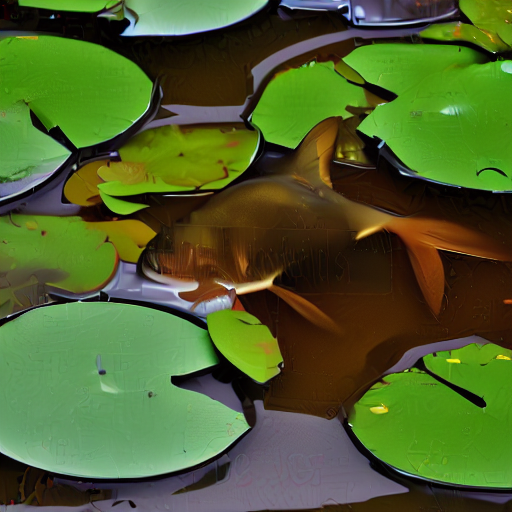}} &
        \adjustbox{valign=m}{\includegraphics[width=0.094\linewidth]{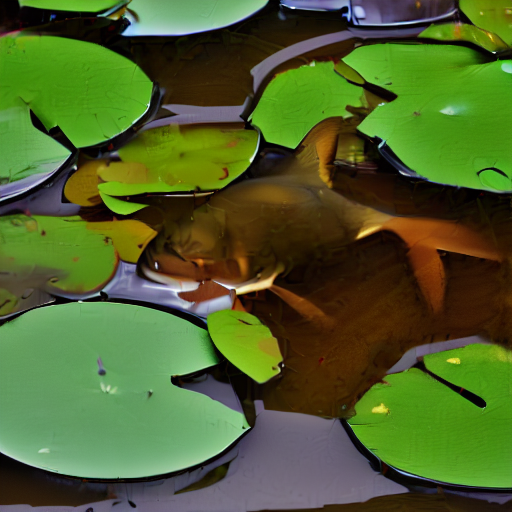}} &
        \adjustbox{valign=m}{\includegraphics[width=0.094\linewidth]{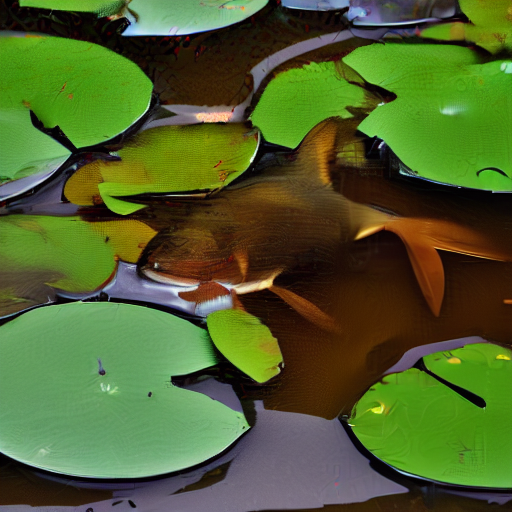}} &
        \adjustbox{valign=m}{\includegraphics[width=0.094\linewidth]{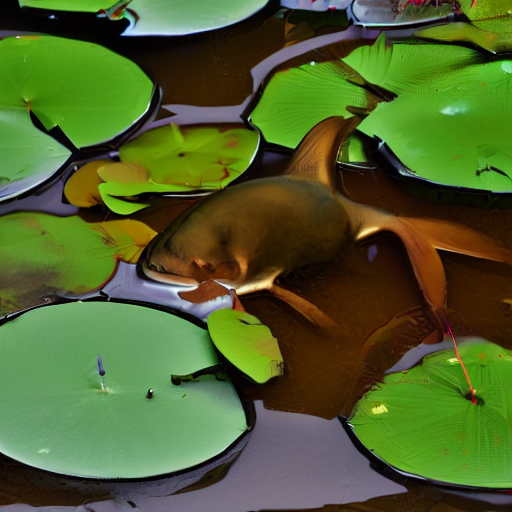}}

        \tabularnewline
        &&&&&&\multicolumn{5}{c}
        {\resizebox{0.48\linewidth}{!}{$\text{PSNR}_{\text{min}}=23.97[\text{dB}]$, $\text{PSNR}_{\text{avg}}=27.19[\text{dB}]$, $\text{PSNR}_{\text{max}} = 28.74[\text{dB}]$}}

    \end{tabular}
    \caption{
    \textbf{Reconstruction from different seeds:}
    Reconstructions from multiple $z_T$ seeds were generated via the Sequential Inversion Block (see \cref{subsec:memory_of_image}) using 10 different support images (left) for two target concepts: Van Gogh and Tench. While preserving each concept’s core appearance, reconstructions vary slightly in background texture, blurriness, and minor elements (\eg, shirt buttons, fins). 
    The average cosine distances between seeds are 0.58 for Van Gogh and 0.69 for Tench.
    These images were generated from an ESD~\cite{gandikota2023esd} model that ablated the Van Gogh (Top) and Tench (Bottom) concepts.}
    \label{fig:image_collage}
\end{figure*}

%% file: figures/18_image_forget_psnr_and_nll_score_bar_plots/figure.tex
\begin{figure*}[h]
    \centering
    \begin{tabular}{cc}

    \resizebox{0.5\linewidth}{!}{{\includegraphics[]{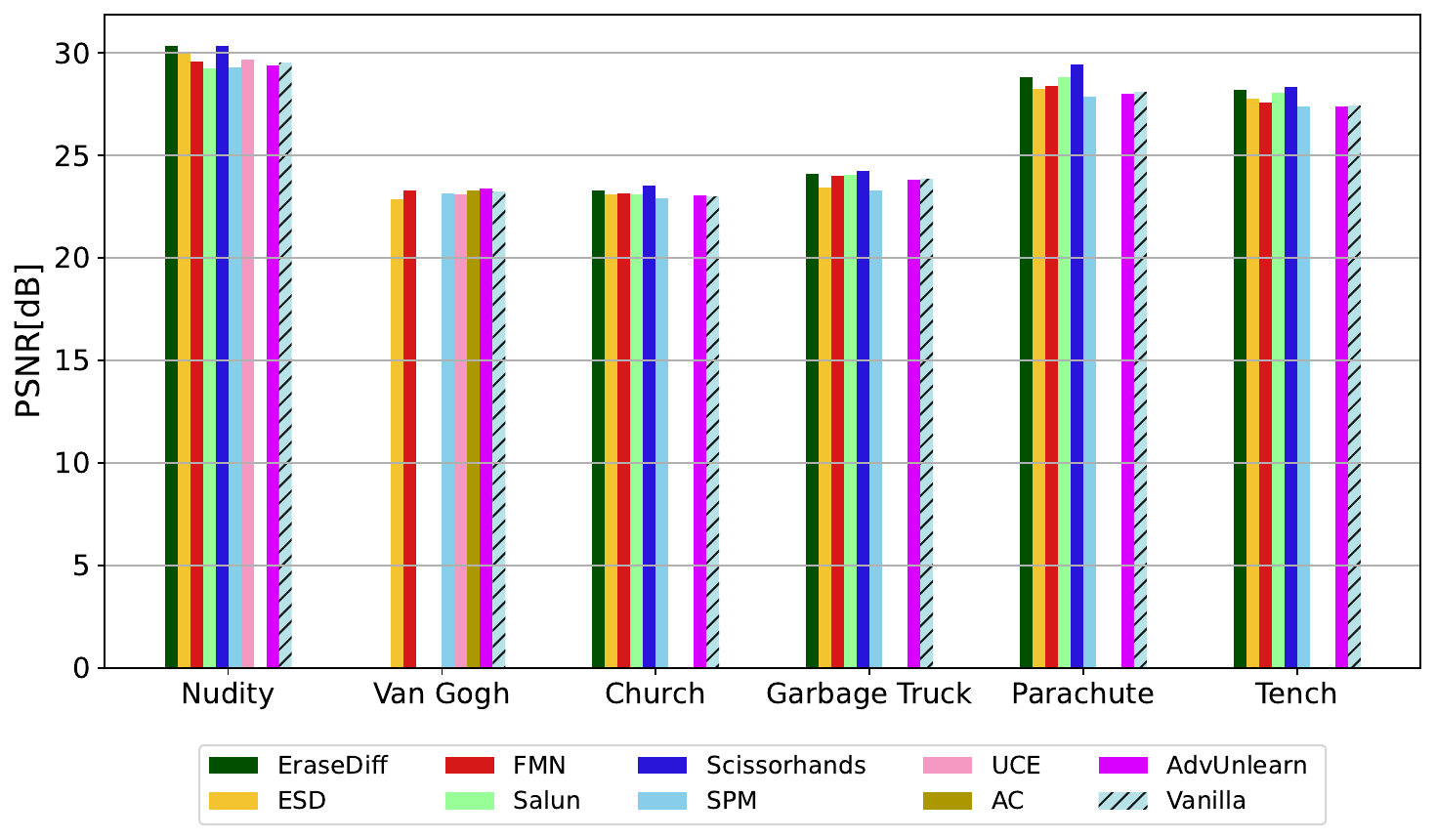}}} &
    \resizebox{0.5\linewidth}{!}{{\includegraphics[]{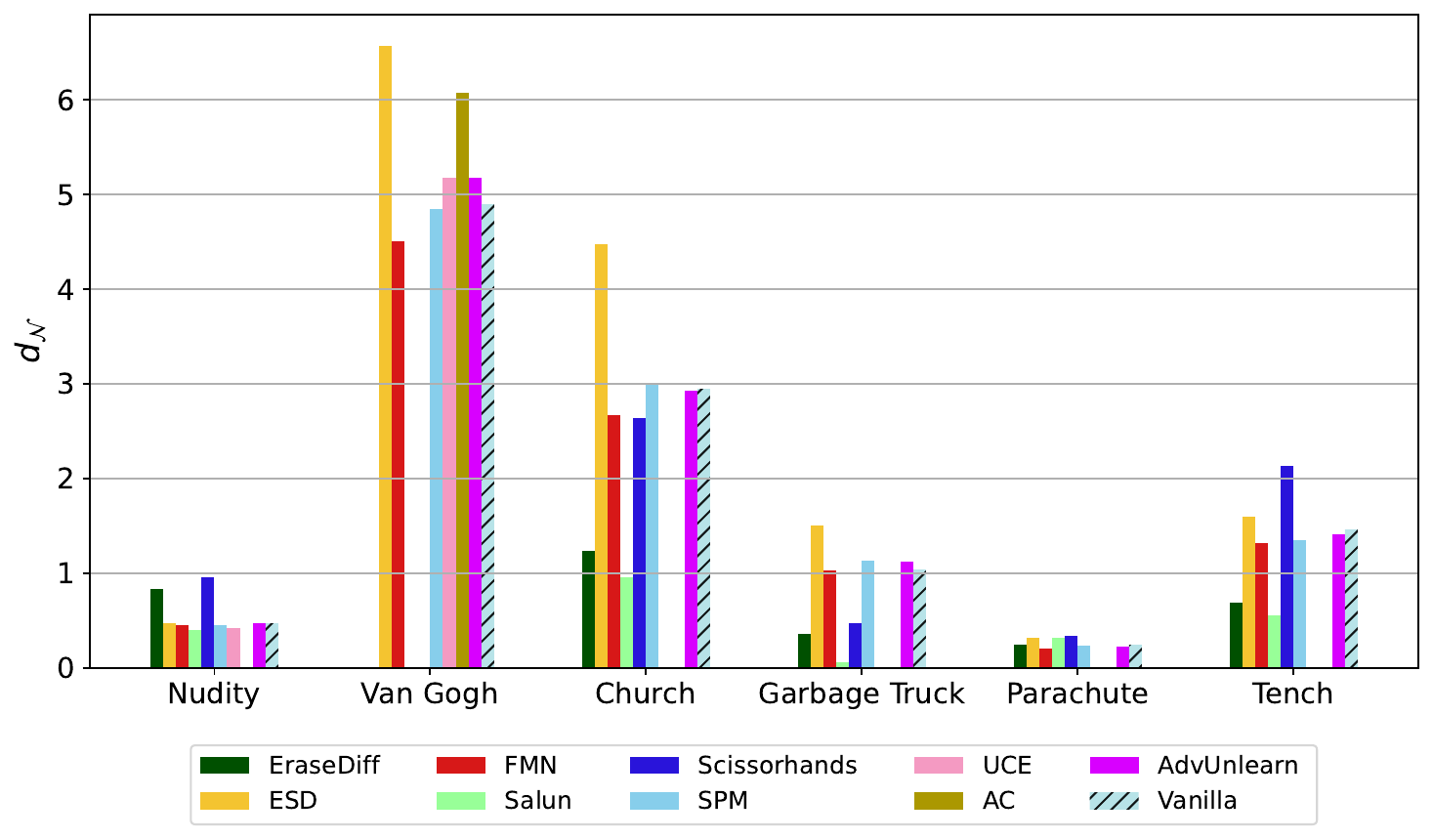}}}\\
    \scriptsize{(a) PSNR [dB] }& \scriptsize{(b) $d_{\mathcal{N}}(E, R)$ }
    \end{tabular}
    \vspace{-3pt}
    \caption{\textbf{Distant latents reconstruct erased images:} We report the mean reconstruction PSNR (a) and our proposed relative distance (b), for different models and concepts (see~\cref{fig:concept-level-results} for more details), 
    obtained using our \textit{sequential inversion block} process resulting in different distant latents for each image.}
    \label{fig:image-level-results}
\end{figure*}

%% file: figures/23_average_cosine_distance/figure.tex
\begin{figure}[h]
    \centering
    
    \resizebox{1\linewidth}{!}{{\includegraphics[]{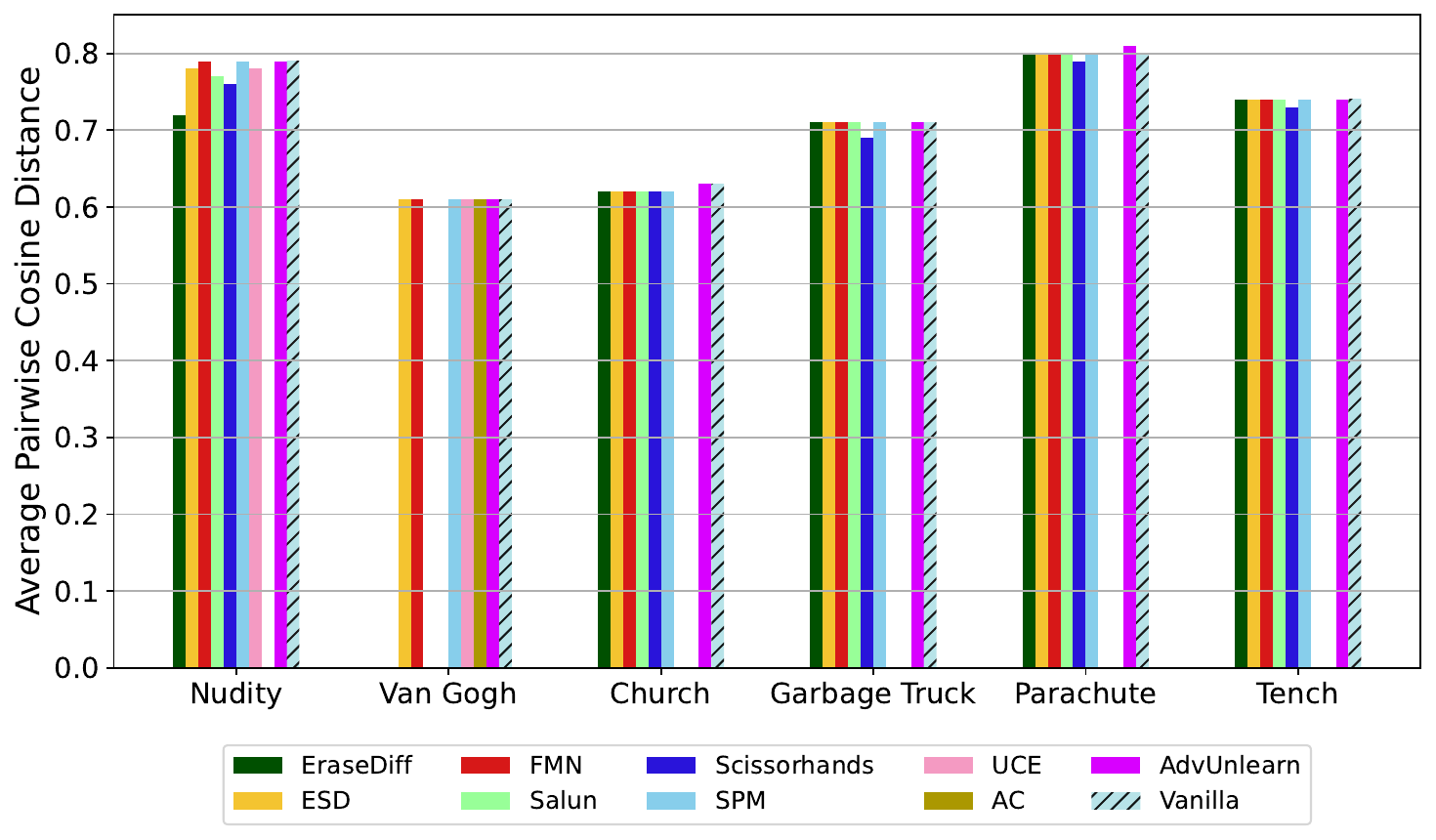}}}
    \caption{\textbf{Average Pairwise Cosine Distance:} For each model that ablated each concept, and for each target image $\mathcal{I}_q$, we average the pairwise cosine distance ($1 - \text{cosine similarity}$) between all the produced $z_T^{(s_i\rightarrow q)}$ seed latents. Then, we average the results over all target images per each model and concept.}
    \label{fig:avg_cos_dist}
\end{figure}

%% file: figures/08_inverting_scrambled_image/figure.tex
\begin{figure}[t]
    \centering
            \setlength{\tabcolsep}{0.5pt} %
        \renewcommand{\arraystretch}{0.6} %
    \begin{tabular}{cccc}
    \begin{subfigure}{0.245\linewidth}
        \centering
        \includegraphics[width=\linewidth]{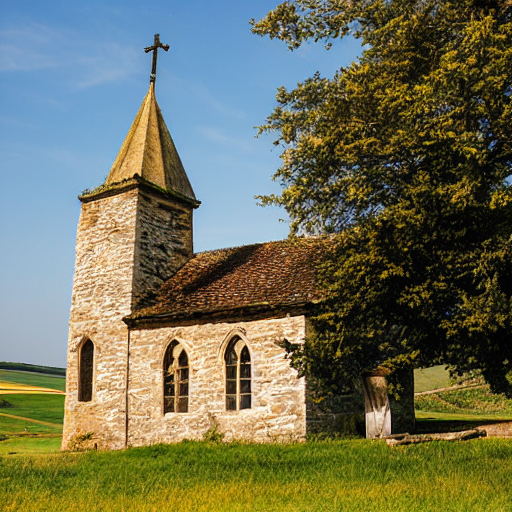}
        
        \caption{$0.3245$}
        \label{fig:inverting_scrambled_image:original}
    \end{subfigure}
        &

            \begin{subfigure}{0.245\linewidth}
        \centering
        \includegraphics[width=\linewidth]{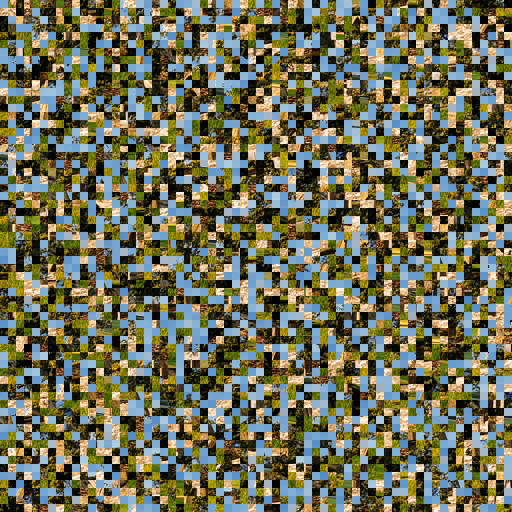}
        
        \caption{$0.1879$}
        \label{fig:inverting_scrambled_image:scrambled}
    \end{subfigure}
    &

        \begin{subfigure}{0.245\linewidth}
        \centering
        \includegraphics[width=\linewidth]{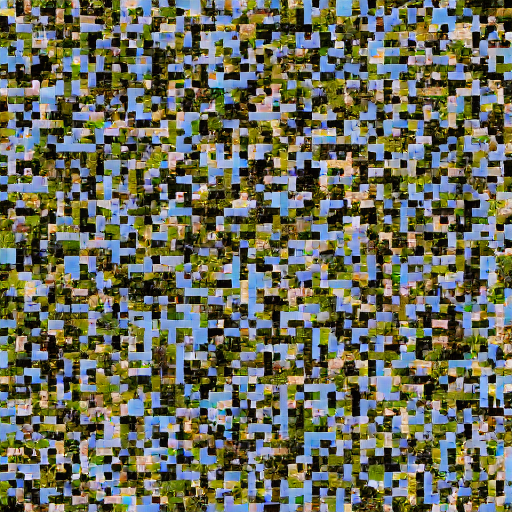}
        
        \caption{$0.1737$}
        \label{fig:inverting_scrambled_image:inverted}
    \end{subfigure}
    &

        \begin{subfigure}{0.245\linewidth}
        \centering
        \includegraphics[width=\linewidth]{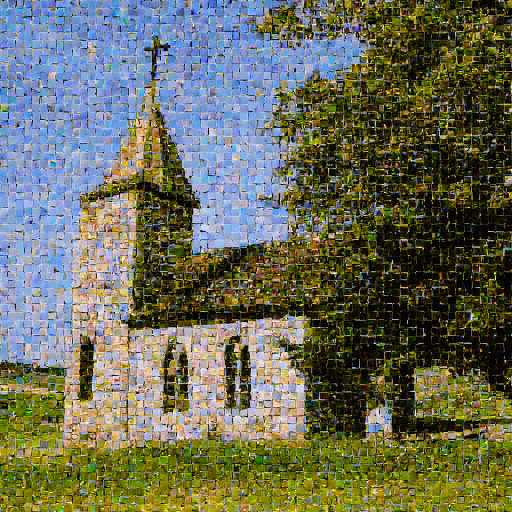}
        
        \caption{$0.2906$}        \label{fig:inverting_scrambled_image:descrambled}
    \end{subfigure}
    \end{tabular}
    \vspace{-5pt}
        \caption{
        \textbf{Ablated models generalize to shuffled images:}
             For a diffusion model that ablates the concept Church, we take a church image in (a), split it to patches of shape $8 \times 8$ and shuffle them to obtain image (b). Then, we invert the image in (b) and regenerate it to obtain the reconstructed image in (c). Finally, we revert the shuffle of patches to obtain the image at (d). Below each image we report the CLIP score of that image \wrt the text ``church".
        }
    \label{fig:inverting_scrambled_image}
\end{figure}

%% file: sections/06_limitations.tex
\section{Limitations}
\label{sec:limitations}
\vspace{-5pt}

The analysis in this paper assumes a white-box setting, assuming access to the model's weights and the ability to inverse it. Although this may limit the generalization of our findings, it enables a controlled exploration of how well-erased concepts can be reconstructed within the model.

In ~\cref{fig:inverting_scrambled_image}, we show that even when an image associated with the ablated concept Church is scrambled, inverted, and then reassembled, the model retains certain associations with the original concept. Starting with a church image (\cref{fig:inverting_scrambled_image:original}), we shuffle its patches to create (\cref{fig:inverting_scrambled_image:scrambled}), invert the scrambled version to produce (\cref{fig:inverting_scrambled_image:inverted}), and then reassemble the patches in (\cref{fig:inverting_scrambled_image:descrambled}). Although the concept classifier score for ``church" drops significantly (from 0.99 to approximately $10^{-4}$), the CLIP similarity to the caption ``church" decreases by only $10\%$.

While models are not expected to ``forget" scrambled versions of erased concepts, this result highlights a significant concern: diffusion models may generalize well even to unusual, pixelated images (such as~\cref{fig:inverting_scrambled_image:scrambled}), successfully inverting them despite their atypical structure.
This generalization ability appears to conflict with concept erasure, as models may inadvertently retain latent representations of ablated concepts. 
While it may seem that inversion is too powerful for this task, we notice that applying our analysis gives coherent results, as the likelihood of the seed that results from the shuffled image is lower, and the PSNR is worse. Specifically, the NLL of the retrieved $z_T$ seed for the shuffled image (\cref{fig:inverting_scrambled_image:scrambled}) is $23.89$K, compared to $23.05$K for the original church image (\cref{fig:inverting_scrambled_image:original}). The reconstruction quality of the recovered noise is also low, with a PSNR score of only $15$dB when compared to (\cref{fig:inverting_scrambled_image:inverted}).

%% file: sections/08_conclusions.tex
\vspace{-5pt}
\section{Conclusions}
\vspace{-5pt}
As diffusion models become more accessible and common to the public, the importance of the safety and privacy of these models increases. 
Recent papers address this concern, developing essential methods for editing diffusion models' outputs to ensure a safer and more controlled generation. 
Previous methods sought to limit the generative capabilities of specific concepts by disrupting the ability to generate these concepts through {\em descriptive text}.
In this work, we hypothesize that an ablated model should not have a high likelihood {\em seed vector} that can be used to generate a high-quality ablated image.
We show, across many methods and different categories, that previous attempts did not truly erase concepts. We do so, by introducing an analysis on the reconstruction quality of images from the erased concepts, and on the likelihood of its corresponding latent seeds. We hope our proposed analysis encourages further research on reliable concept erasure evaluation.

%% file: appendix.tex
\appendix
\arxiv{\clearpage}{}
\arxiv{\section*{\LARGE Appendices}}{}

\arxiv{}{\setcounter{figure}{10}}
\arxiv{}{\setcounter{equation}{7}}

In the next sections, we provide additional details and results that further support our analysis and offer a more comprehensive understanding of the findings we presented in the main paper.
\cref{supp:sec:likelihood_effect} provides additional information on how the likelihood of latents affects image generation. \cref{supp:sec:different_init} explains our initialization choice when searching for distant latents that can generate a given query image $I_q$. \cref{supp:sec:further_analysis} provides additional results for the experiments shown in the paper, including metrics that were not discussed, such as CLIP-score and a concept detector accuracy. \cref{supp:sec:nll_analysis} contains an analysis of the distribution of the NLL of multivariate normally distribute vectors. \cref{supp:sec:inversion_params} contains results that justify our choice of inversion method and its parameters.

\input{supp_sections/low_likely_images}

\input{supp_sections/different_initializations}

\input{supp_sections/further_analysis}

\input{supp_sections/0999_distribution_of_nlls}

\input{supp_sections/inversion_parameters}

%% file: supp_sections/low_likely_images.tex
\section{Likelihood effect on generation}
\label{supp:sec:likelihood_effect}
\input{figures/06_inversion_of_unlikely/figure}

In this section, we focus on further examining the effect of the likelihood of $z_T$ on the generated image by a given diffusion model. As explained in~\arxiv{\cref{sec:limitations}}{Sec.~4}, inversion is a powerful tool that can be used to generate images with different likelihoods. But, examining these generated images along with reconstruction error can give more information. For example, in~\cref{supp:fig:inversion_unlikely}, we see that while inversion is used to transform a {\em very unlikely} image to an image with reasonable likelihood, the reconstruction PSNR of this process is poor.
\input{figures/05_likelihood_change/figure}

We are also interested in the relation between likelihood to generation quality. As shown on~\cref{suppfig:likelihood_change}, while the $\Vec{0}$ vector has the lowest (best) NLL, its generation quality is poor. This is due to the fact that the model was trained using random samples from the standard normal distribution, and (with high probability) have not been given the $\Vec{0}$ as input for generation. This coincides with the work of Samuel \etal~\cite{samuel2024norm_guided_latent}, demonstrating that diffusion models are learned using latents with a specific norm range. This conclusion is important for our analysis, as we do not use the NLL as an absolute score, but rather as a relative score compared to the NLL of the standard normal distribution (see ~\cref{eq:relative_similarity}).

%% file: figures/06_inversion_of_unlikely/figure.tex
\begin{figure}[h]
    \centering
        \setlength{\tabcolsep}{4.5pt} %
        \renewcommand{\arraystretch}{0.6} %
    
    \begin{tabular}{@{}cc@{}}
   \includegraphics[width=0.47\linewidth]{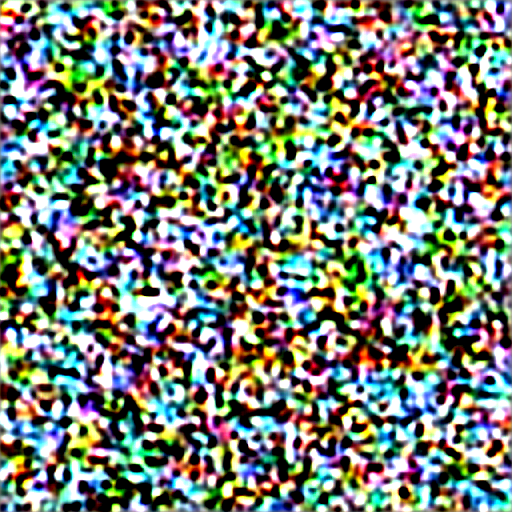} 
   
    &
    \includegraphics[width=0.47\linewidth]{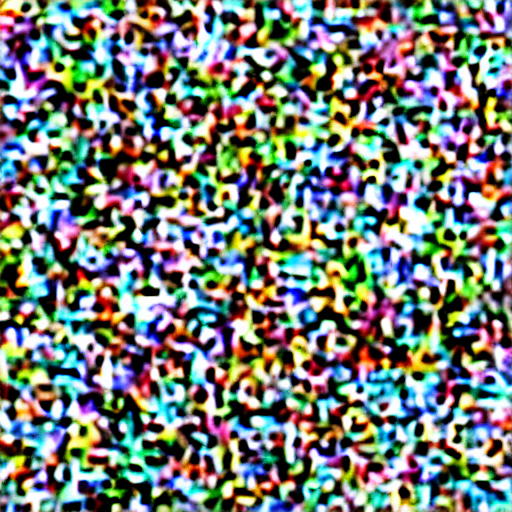}
    \tabularnewline

    \small{NLL $\approx8\cdot10^7$}&\small{NLL $\approx24$K}
    
    \end{tabular}
    
        \caption{\textbf{Inversion of low likelihood images.} A low likelihood latent can be used to generate an image (left). The image can be inverted to find a latent that generates a similar image (right), with PSNR=$19.64 [ \text{dB} ]$.}
    \label{supp:fig:inversion_unlikely}
\end{figure}

%% file: figures/05_likelihood_change/figure.tex
\begin{figure}[t]
    \centering
        \setlength{\tabcolsep}{0.5pt} %
        \renewcommand{\arraystretch}{0.6} %
    \begin{tabular}{@{}p{0.05\linewidth}ccccc@{}}
      \toprule
      \scriptsize{NLL}&\scriptsize{$15$K}&\scriptsize{$20$K}&\scriptsize{$23.2$K}&\scriptsize{$26.3$K}&\scriptsize{$31.2$K}
      \tabularnewline
      \multicolumn{1}{c}{\scriptsize{$\alpha$}}&\scriptsize{$0$}&\scriptsize{$100$}&\scriptsize{$128$}&\scriptsize{$150$}&\scriptsize{$180$}
      \tabularnewline

      \midrule
      &\includegraphics[width=0.18\linewidth]{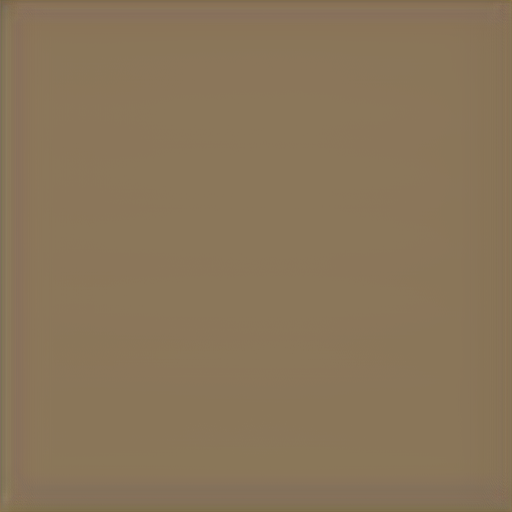} & \includegraphics[width=0.18\linewidth]{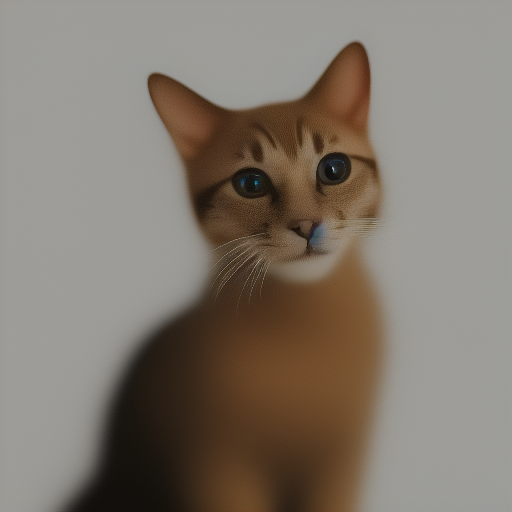} & 
      \includegraphics[width=0.18\linewidth]{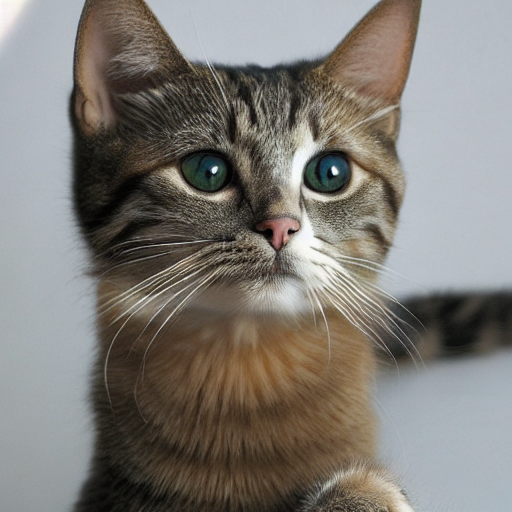} & 
      \includegraphics[width=0.18\linewidth]{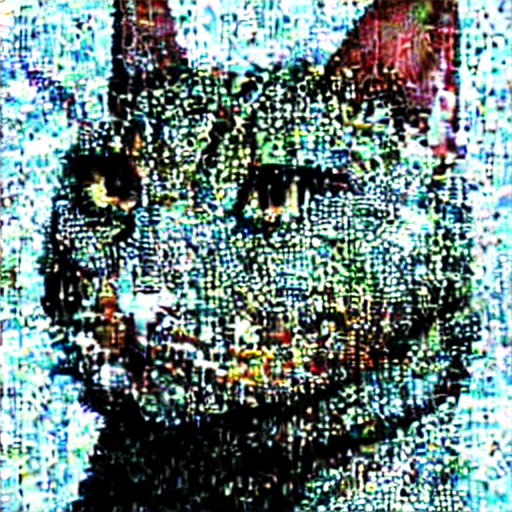} & 
      \includegraphics[width=0.18\linewidth]{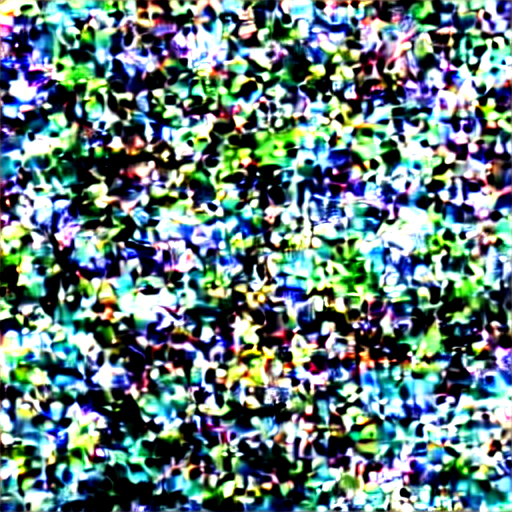} \tabularnewline
      
      \bottomrule
    \end{tabular}
    \renewcommand{\arraystretch}{1} %
        \caption{\textbf{Generation using latents with varying likelihoods:} Using the same caption ``cat'', the likelihood of the initial latent seed controls the generation quality. This is done by sampling $z\sim \mathcal{N}(0, 1)$ and applying $Y = \alpha \cdot \frac{z}{||z||}$ (\ie, using the same vector with a scaled norm of $\alpha$).}
    \label{suppfig:likelihood_change}
\end{figure}

%% file: supp_sections/different_initializations.tex
\input{figures/27_diff_init_sample_close/figure}

\input{figures/28_diff_init_sample_far/figure}
\input{figures/29_diff_init_sib_random_noise/figure}

\section{Different initializations for distant memories retrieval}
\label{supp:sec:different_init}

In~\arxiv{\cref{subsec:memory_of_image}}{Sec.~3.3}, we suggest applying our sequential inversion block (SIB), starting from arbitrary support images to retrieve distant memories of a given ablated target image. Next, we present a few straightforward alternatives and discuss their drawbacks. 

Instead of performing a VAE decoder inversion, one could suggest utilizing the encoder of the VAE. Given an image $I_q$, the encoder returns parameters for a normal distribution, \ie, $\text{Enc}(I_q) = \mathcal{N}(\mu_{I_q}, \Sigma_{I_q})$. The distribution can be used to sample multiple different latents, in close proximity. In ~\arxiv{\cref{sec:analysis}}{Sec.~3}, we do not sample multiple latents, but rather we use a latent $z_0$ which is the mean of the distribution, $\mu_{I_q}$. 
\cref{fig:diff_init_sample_close} shows the PSNR and distances results for these latents. The reconstruction quality is high, but all memories turn out the same, displaying an average pairwise cosine distance of $0$. 

In~\cref{fig:diff_init_sample_far}, in order to examine the case of more distant latents, we sample from $\mathcal{N}(\mu_{I_q}, \Sigma_{I_q})$ but add a standard normal random noise (normalized across its channels dimension) and scaled by a factor of 10. As can be seen, the average pairwise cosine distance is much higher and resembles the average pairwise cosine distance presented by our solution in~\cref{fig:avg_cos_dist}. However, the PSNR is considerably lower, suggesting that this method did not reconstruct images that resemble $I_q$. 

Finally, we show in~\cref{fig:diff_init_sib_random_noise}, that applying SIB, starting from randomly sampled latents, is also suboptimal. Although the reconstruction quality is sufficient, the average pairwise cosine distance is lower than our suggested method (See~\arxiv{\cref{fig:avg_cos_dist}}{Fig.~9}). 

We conclude that trivial random initializations are suboptimal, and are inferior compared to our method in~\arxiv{\cref{subsec:memory_of_image}}{Sec.~3.3}.

%% file: figures/27_diff_init_sample_close/figure.tex
\begin{figure*}[h]

\begin{tikzpicture}
    \node (img1) at (0, 0) {\includegraphics[width=0.49\textwidth]{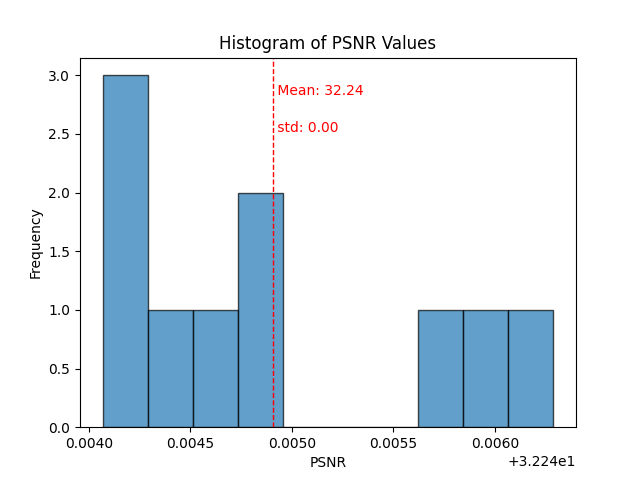} };
    \node[below=0.48cm of img1] {\small{(a) PSNR [dB]}};
    
    \node (img2) at (9, 0) {\includegraphics[width=0.49\textwidth]{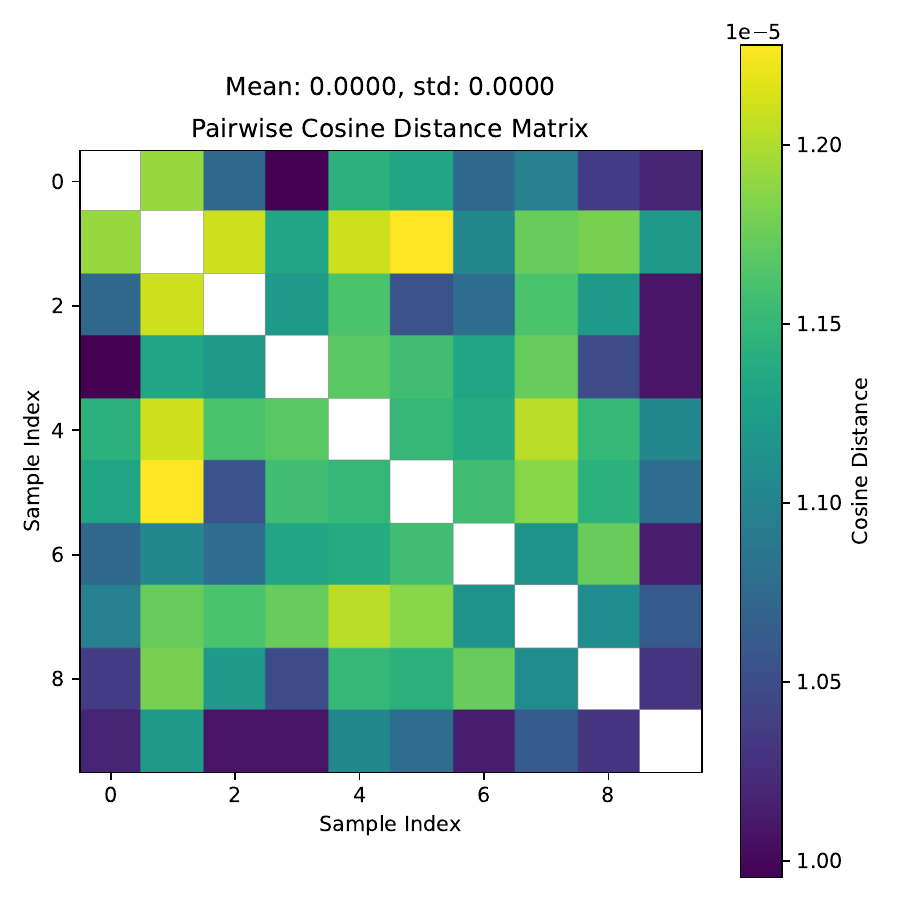}};
    \node[below=-0.6cm of img2] {\small{(b) Pairwise cosine distance.}};
\end{tikzpicture}
    
        \caption{\textbf{Sample near:} randomly sample 10 latents from $\mathcal{N}(\mu = \text{Enc}(I_q), \Sigma_{I_q})$.}

    \label{fig:diff_init_sample_close}
\end{figure*}

%% file: figures/28_diff_init_sample_far/figure.tex
\begin{figure*}[h]

\begin{tikzpicture}
    \node (img1) at (0, 0) {\includegraphics[width=0.49\textwidth]{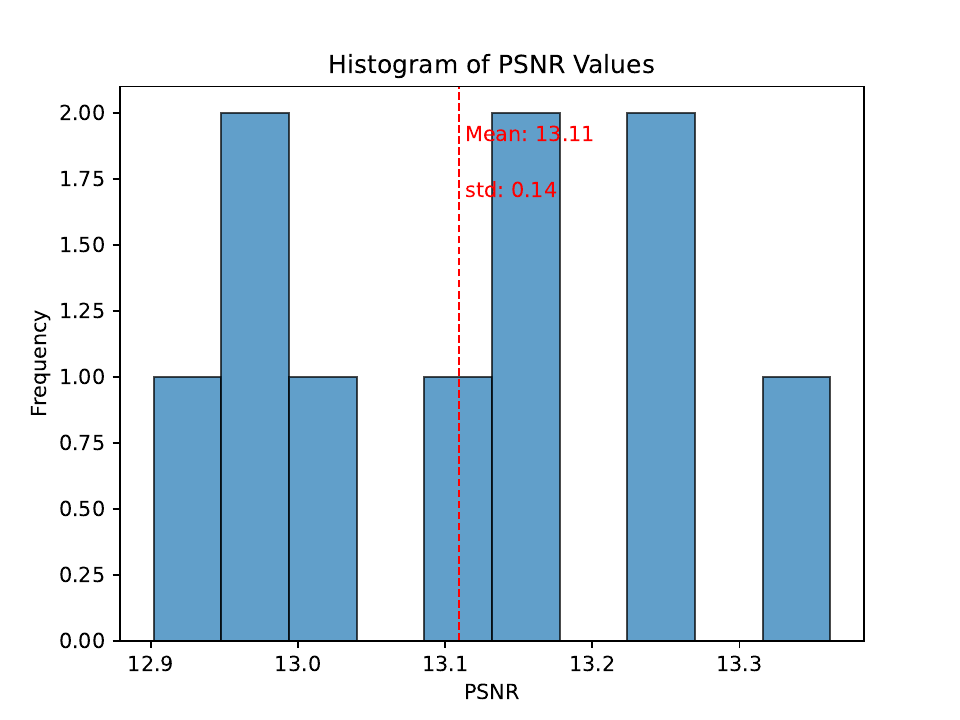} };
    \node[below=0.48cm of img1] {\small{(a) PSNR [dB]}};
    
    \node (img2) at (9, 0) {\includegraphics[width=0.49\textwidth]{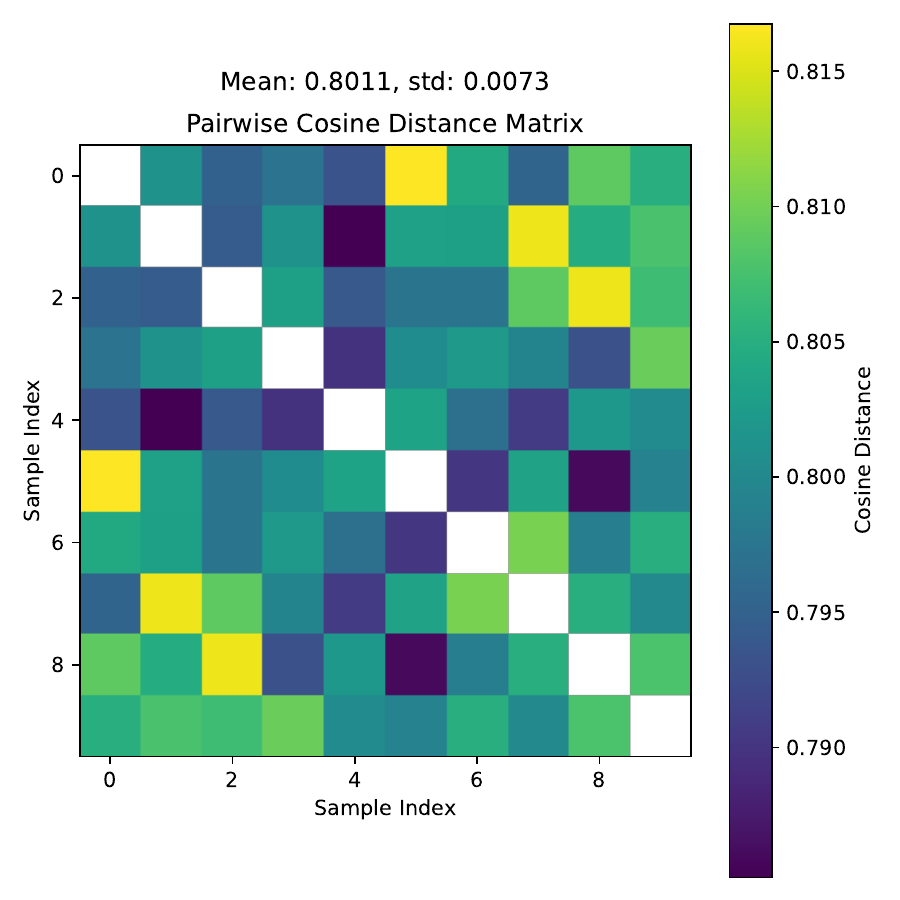}};
    \node[below=-0.6cm of img2] {\small{(b) Pairwise cosine distance.}};
\end{tikzpicture}

        \caption{\textbf{Sample far:} randomly sample 10 latents from $\mathcal{N}(\mu = \text{Enc}(I_q), \Sigma_{I_q})$. For each sample, add a random noise.}
    \label{fig:diff_init_sample_far}
\end{figure*}

%% file: figures/29_diff_init_sib_random_noise/figure.tex
\begin{figure*}[h]

\begin{tikzpicture}
    \node (img1) at (0, 0) {\includegraphics[width=0.49\textwidth]{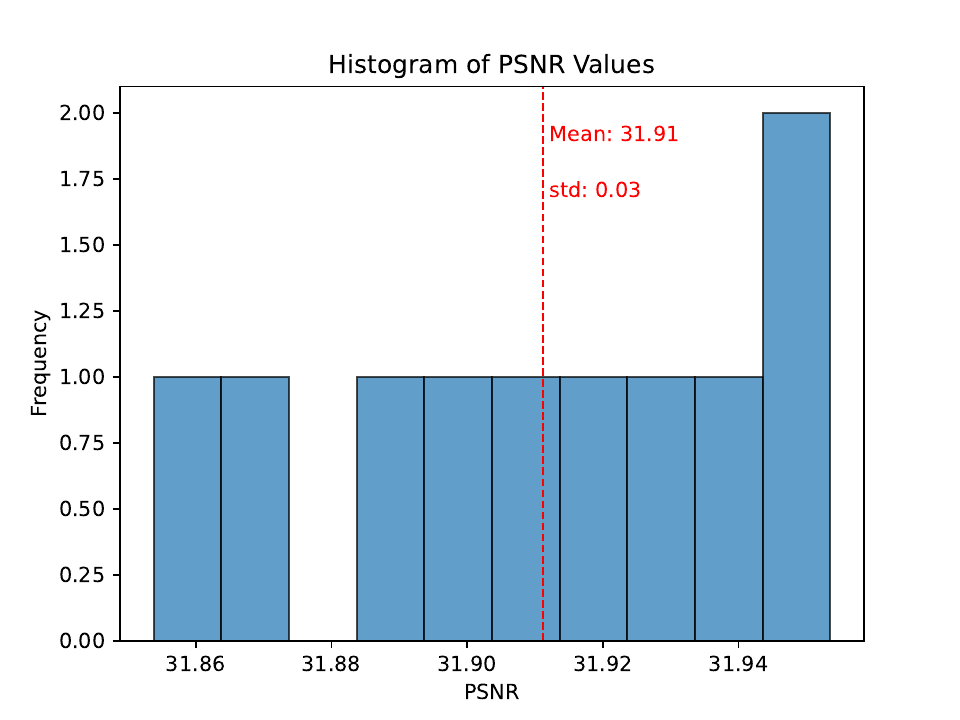} };
    \node[below=0.48cm of img1] {\small{(a) PSNR [dB]}};
    
    \node (img2) at (9, 0) {\includegraphics[width=0.49\textwidth]{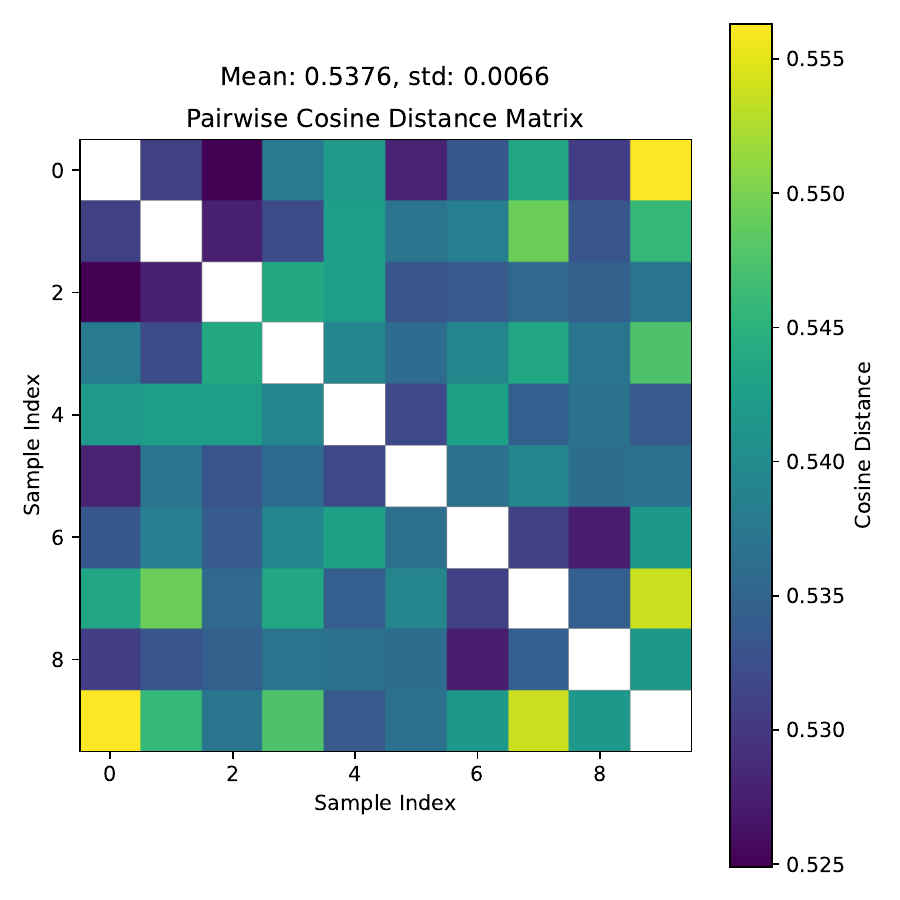}};
    \node[below=-0.6cm of img2] {\small{(b) Pairwise cosine distance.}};
\end{tikzpicture}

        \caption{\textbf{SIB on random noise:} randomly sample 10 initializations for SIB.}
    \label{fig:diff_init_sib_random_noise}
\end{figure*}

%% file: supp_sections/further_analysis.tex
\section{Further analysis}
\label{supp:sec:further_analysis}
Next, we present additional analysis regarding the experiments in~\arxiv{\cref{sec:analysis}}{Sec.~3}. Specifically, ~\cref{supp:tab:concept_nudity,supp:tab:concept_church,supp:tab:concept_parachute,supp:tab:concept_tench,supp:tab:concept_garbage_truck,supp:tab:concept_vangogh} and~\cref{supp:tab:image_nudity,supp:tab:image_church,supp:tab:image_parachute,supp:tab:image_tench,supp:tab:image_garbage_truck,supp:tab:image_vangogh} contain extended results for the experiments detailed in~\arxiv{\cref{subsec:memory_of_concept,subsec:memory_of_image}}{Secs.~3.2 and 3.3} (and visualized in~\arxiv{\cref{fig:concept-level-results,fig:image-level-results}}{Figs.~5 and~8}), respectively.
Each table contains results that correspond to one concept. These tables contain the scores discussed in the main paper, \ie, PSNR and $d_\mathcal{N}(E, R)$, along with a concept classifier detection score, CLIP-score and the EMD between different distributions. The presented EMD results are:
\begin{enumerate}
    \item \label{supp:d_n_numerator} $E, \mathcal{N}$ --- The EMD between the NLL of latents in the erased set $E$ and the NLL of standard normal samples, \ie, $\text{EMD}\left(\nllzt(E), NLL(\mathcal{N}) \right)$.
    \item \label{supp:d_n_denominator} $R, \mathcal{N}$ --- The same as above, using latents from the reference set $R$, \ie, $\text{EMD}\left(\nllzt(R), NLL(\mathcal{N})\right)$.
    \item $E, R$ --- The EMD between latents in the erased and reference sets, \ie, $\text{EMD}\left(\nllzt(E), \nllzt(R)\right)$.
\end{enumerate}
\cref{supp:d_n_numerator,supp:d_n_denominator} serve as the numerator and denominator of $d_\mathcal{N}(\cdot, \cdot)$ (see~\arxiv{\cref{eq:relative_similarity}}{Eq.~(5)}), respectively. 
 \cref{supp:tab:image_nudity,supp:tab:image_church,supp:tab:image_parachute,supp:tab:image_tench,supp:tab:image_garbage_truck,supp:tab:image_vangogh} also contain the average distances between all $z^{(s_i \rightarrow q)}_T$ and $z^q_T$ (see~\arxiv{\cref{fig:avg_cos_dist}}{Fig.~9}). These values are shown in both euclidean distance and cosine similarity.

\input{tables/concept_forget/nudity_church_combined}

\input{tables/concept_forget/parachute_tench_combined}

\input{tables/concept_forget/garbage_truck_van_gogh_combined}

\input{tables/image_forget/nudity}

\input{tables/image_forget/church}

\input{tables/image_forget/parachute}

\input{tables/image_forget/tench}

\input{tables/image_forget/garbage_truck}

\input{tables/image_forget/vangogh}

%% file: tables/concept_forget/nudity_church_combined.tex
\newcolumntype{C}[1]{>{\centering\arraybackslash}p{#1}}

\begin{table*}[h]
\begin{minipage}[t]{0.49\textwidth}
\label{tab:concept_nudity}

\resizebox{\linewidth}{!}{
\begin{tabular}{l@{} 
C{0.1\linewidth}
C{0.12\linewidth}
C{0.12\linewidth}
C{0.12\linewidth}
C{0.12\linewidth}
C{0.12\linewidth}
C{0.1\linewidth}
}

\toprule
&&&\multicolumn{3}{c}{{\small $\text{EMD}$}}
\tabularnewline
\cmidrule(l{2pt}r{2pt}){4-6}

&{\small Detection (\%)} &{ \small  PSNR [dB] } &\multirow{2}{*}{ {\small $E, \mathcal{N}$ }}& \multirow{2}{*}{{\small $R,\mathcal{N}$ }}&
\multirow{2}{*}{{\small $E,R$}}&

\multirow{2}{*}{{ \small $d_{\mathcal{N}}(E, R)$ }}&{\small CLIP-Score }
\tabularnewline
\midrule

EraseDiff~\cite{wu2024erasediff} & 100 & 34.13 & 1565.2K & 2896.0K & 269.1K & 0.54 & 0.31 \tabularnewline
ESD~\cite{gandikota2023esd} & 100 & 34.24 & 281.7K & 194.9K & 15.6K & 1.45 & 0.31 \tabularnewline
FMN~\cite{zhang2024fmn} & 100 & 34.23 & 249.0K & 185.9K & 14.4K & 1.34 & 0.31 \tabularnewline
Salun~\cite{fan2023salun} & 100 & 34.05 & 408.5K & 294.4K & 13.9K & 1.39 & 0.31 \tabularnewline
Scissorhands~\cite{wu2024scissorhands} & 100 & 34.32 & 2090.3K & 2603.4K & 67.1K & 0.80 & 0.32 \tabularnewline
SPM~\cite{lyu2024spm} & 100 & 34.22 & 257.3K & 182.3K & 14.0K & 1.41 & 0.31 \tabularnewline
UCE~\cite{gandikota2024uce} & 100 & 34.22 & 263.8K & 192.6K & 13.3K & 1.37 & 0.31 \tabularnewline
AdvUnlearn~\cite{zhang2024advunlearn} & 100 & 34.22 & 262.2K & 188.7K & 13.9K & 1.39 & 0.31 \tabularnewline
Vanilla~\cite{rombach2022stable_diffusion} & 100 & 34.21 & 287.0K & 210.1K & 14.0K & 1.37 & 0.31 \tabularnewline

\bottomrule
\end{tabular}}
\caption{\textbf{Ablated concept: Nudity.}} 

\label{supp:tab:concept_nudity}

\end{minipage}
\hspace{0.02\linewidth}
\begin{minipage}[t]{0.49\linewidth}
\label{tab:concept_church}
\resizebox{\linewidth}{!}{
\begin{tabular}{l@{} 
C{0.1\linewidth}
C{0.12\linewidth}
C{0.12\linewidth}
C{0.12\linewidth}
C{0.12\linewidth}
C{0.12\linewidth}
C{0.1\linewidth}
}

\toprule
&&&\multicolumn{3}{c}{{\small $\text{EMD}$}}
\tabularnewline
\cmidrule(l{2pt}r{2pt}){4-6}

&{\small Detection (\%)} &{ \small  PSNR [dB] } &\multirow{2}{*}{ {\small $E, \mathcal{N}$ }}& \multirow{2}{*}{{\small $R,\mathcal{N}$ }}&
\multirow{2}{*}{{\small $E,R$}}&

\multirow{2}{*}{{ \small $d_{\mathcal{N}}(E, R)$ }}&{\small CLIP-Score }
\tabularnewline
\midrule

EraseDiff~\cite{wu2024erasediff} & 96 & 28.73 & 434.5K & 229.8K & 34.2K & 1.89 & 0.31 \tabularnewline
ESD~\cite{gandikota2023esd} & 96 & 28.76 & 347.7K & 173.9K & 33.7K & 2.00 & 0.31 \tabularnewline
FMN~\cite{zhang2024fmn} & 96 & 28.75 & 436.4K & 223.2K & 37.0K & 1.96 & 0.31 \tabularnewline
Salun~\cite{fan2023salun} & 96 & 28.74 & 568.1K & 318.5K & 38.5K & 1.78 & 0.31 \tabularnewline
Scissorhands~\cite{wu2024scissorhands} & 96 & 28.71 & 402.9K & 185.5K & 42.9K & 2.17 & 0.31 \tabularnewline
SPM~\cite{lyu2024spm} & 96 & 28.74 & 413.7K & 205.9K & 37.1K & 2.01 & 0.31 \tabularnewline
AdvUnlearn~\cite{zhang2024advunlearn} & 96 & 28.71 & 433.5K & 214.4K & 39.4K & 2.02 & 0.31 \tabularnewline
Vanilla~\cite{rombach2022stable_diffusion} & 96 & 28.71 & 413.7K & 201.0K & 39.4K & 2.06 & 0.31 \tabularnewline

\bottomrule
\end{tabular}}
\caption{\textbf{Ablated concept: Church.}}

\label{supp:tab:concept_church}
\end{minipage}
\end{table*}

%% file: tables/concept_forget/parachute_tench_combined.tex
\newcolumntype{C}[1]{>{\centering\arraybackslash}p{#1}}

\begin{table*}[h]
\begin{minipage}[t]{0.49\linewidth}
\label{tab:concept_parachute}
\resizebox{\linewidth}{!}{

\begin{tabular}{l@{} 
C{0.1\linewidth}
C{0.12\linewidth}
C{0.12\linewidth}
C{0.12\linewidth}
C{0.12\linewidth}
C{0.12\linewidth}
C{0.1\linewidth}
}

\toprule
&&&\multicolumn{3}{c}{{\small $\text{EMD}$}}
\tabularnewline
\cmidrule(l{2pt}r{2pt}){4-6}

&{\small Detection (\%)} &{ \small  PSNR [dB] } &\multirow{2}{*}{ {\small $E, \mathcal{N}$ }}& \multirow{2}{*}{{\small $R,\mathcal{N}$ }}&
\multirow{2}{*}{{\small $E,R$}}&

\multirow{2}{*}{{ \small $d_{\mathcal{N}}(E, R)$ }}&{\small CLIP-Score }
\tabularnewline
\midrule

EraseDiff~\cite{wu2024erasediff} & 96 & 32.44 & 521.6K & 256.2K & 47.3K & 2.04 & 0.32 \tabularnewline
ESD~\cite{gandikota2023esd} & 98 & 32.47 & 386.0K & 159.1K & 49.5K & 2.43 & 0.32 \tabularnewline
FMN~\cite{zhang2024fmn} & 96 & 32.33 & 507.6K & 238.9K & 50.5K & 2.12 & 0.32 \tabularnewline
Salun~\cite{fan2023salun} & 96 & 32.50 & 631.1K & 356.7K & 38.9K & 1.77 & 0.32 \tabularnewline
Scissorhands~\cite{wu2024scissorhands} & 94 & 32.48 & 484.0K & 217.7K & 52.5K & 2.22 & 0.32 \tabularnewline
SPM~\cite{lyu2024spm} & 96 & 32.46 & 436.9K & 201.4K & 45.5K & 2.17 & 0.32 \tabularnewline
AdvUnlearn~\cite{zhang2024advunlearn} & 96 & 32.47 & 461.7K & 214.9K & 47.1K & 2.15 & 0.32 \tabularnewline
Vanilla~\cite{rombach2022stable_diffusion} & 96 & 32.47 & 453.1K & 209.1K & 47.0K & 2.17 & 0.32 \tabularnewline

\bottomrule
\end{tabular}}
\caption{\textbf{Ablated concept: Parachute.}}

\label{supp:tab:concept_parachute}

\end{minipage}
\hspace{0.02\linewidth}
\begin{minipage}[t]{0.49\linewidth}
\label{tab:concept_tench}
\resizebox{\linewidth}{!}{

\begin{tabular}{l@{} 
C{0.1\linewidth}
C{0.12\linewidth}
C{0.12\linewidth}
C{0.12\linewidth}
C{0.12\linewidth}
C{0.12\linewidth}
C{0.1\linewidth}
}

\toprule
&&&\multicolumn{3}{c}{{\small $\text{EMD}$}}
\tabularnewline
\cmidrule(l{2pt}r{2pt}){4-6}

&{\small Detection (\%)} &{ \small  PSNR [dB] } &\multirow{2}{*}{ {\small $E, \mathcal{N}$ }}& \multirow{2}{*}{{\small $R,\mathcal{N}$ }}&
\multirow{2}{*}{{\small $E,R$}}&

\multirow{2}{*}{{ \small $d_{\mathcal{N}}(E, R)$ }}&{\small CLIP-Score }
\tabularnewline
\midrule

EraseDiff~\cite{wu2024erasediff} & 76 & 31.41 & 125.6K & 269.0K & 41.4K & 0.47 & 0.31 \tabularnewline
ESD~\cite{gandikota2023esd} & 78 & 31.41 & 86.5K & 206.7K & 53.1K & 0.42 & 0.31 \tabularnewline
FMN~\cite{zhang2024fmn} & 78 & 31.41 & 130.1K & 245.2K & 27.7K & 0.53 & 0.31 \tabularnewline
Salun~\cite{fan2023salun} & 76 & 31.41 & 167.2K & 341.2K & 40.8K & 0.49 & 0.31 \tabularnewline
Scissorhands~\cite{wu2024scissorhands} & 76 & 31.41 & 264.7K & 256.4K & 8.1K & 1.03 & 0.31 \tabularnewline
SPM~\cite{lyu2024spm} & 78 & 31.41 & 108.4K & 214.8K & 29.4K & 0.50 & 0.31 \tabularnewline
AdvUnlearn~\cite{zhang2024advunlearn} & 78 & 31.41 & 100.1K & 194.0K & 26.1K & 0.52 & 0.31 \tabularnewline
Vanilla~\cite{rombach2022stable_diffusion} & 78 & 31.41 & 111.0K & 212.7K & 26.1K & 0.52 & 0.31 \tabularnewline
\bottomrule
\end{tabular}}
\caption{\textbf{Ablated concept: Tench.}}
\label{supp:tab:concept_tench}
\end{minipage}
\end{table*}

%% file: tables/concept_forget/garbage_truck_van_gogh_combined.tex
\newcolumntype{C}[1]{>{\centering\arraybackslash}p{#1}}

\begin{table*}[ht]
\begin{minipage}[t]{0.49\linewidth}
    
\label{tab:concept_garbage_truck}
\resizebox{\linewidth}{!}{
\begin{tabular}{l@{} 
C{0.1\linewidth}
C{0.12\linewidth}
C{0.12\linewidth}
C{0.12\linewidth}
C{0.12\linewidth}
C{0.12\linewidth}
C{0.1\linewidth}
}

\toprule
&&&\multicolumn{3}{c}{{\small $\text{EMD}$}}
\tabularnewline
\cmidrule(l{2pt}r{2pt}){4-6}

&{\small Detection (\%)} &{ \small  PSNR [dB] } &\multirow{2}{*}{ {\small $E, \mathcal{N}$ }}& \multirow{2}{*}{{\small $R,\mathcal{N}$ }}&
\multirow{2}{*}{{\small $E,R$}}&

\multirow{2}{*}{{ \small $d_{\mathcal{N}}(E, R)$ }}&{\small CLIP-Score }
\tabularnewline
\midrule

EraseDiff~\cite{wu2024erasediff} & 80 & 28.98 & 447.3K & 261.1K & 25.0K & 1.71 & 0.30 \tabularnewline
ESD~\cite{gandikota2023esd} & 84 & 28.93 & 369.9K & 199.3K & 26.3K & 1.86 & 0.30 \tabularnewline
FMN~\cite{zhang2024fmn} & 82 & 29.00 & 447.0K & 244.1K & 30.5K & 1.83 & 0.30 \tabularnewline
Salun~\cite{fan2023salun} & 82 & 28.98 & 618.6K & 362.4K & 34.1K & 1.71 & 0.30 \tabularnewline
Scissorhands~\cite{wu2024scissorhands} & 86 & 29.01 & 298.0K & 283.1K & 11.2K & 1.05 & 0.30 \tabularnewline
SPM~\cite{lyu2024spm} & 82 & 28.95 & 374.0K & 193.2K & 29.6K & 1.94 & 0.30 \tabularnewline
AdvUnlearn~\cite{zhang2024advunlearn} & 82 & 28.97 & 428.4K & 216.5K & 35.8K & 1.98 & 0.30 \tabularnewline
Vanilla~\cite{rombach2022stable_diffusion} & 82 & 28.97 & 421.7K & 212.7K & 35.4K & 1.98 & 0.30 \tabularnewline
\bottomrule
\end{tabular}}
\caption{\textbf{Ablated concept: Garbage Truck.}}

\label{supp:tab:concept_garbage_truck}
\end{minipage}
\hspace{0.02\linewidth}
\begin{minipage}[t]{0.49\linewidth}
    \label{tab:concept_vangogh}
\resizebox{\linewidth}{!}{
\begin{tabular}{l@{} 
C{0.1\linewidth}
C{0.12\linewidth}
C{0.12\linewidth}
C{0.12\linewidth}
C{0.12\linewidth}
C{0.12\linewidth}
C{0.1\linewidth}
}

\toprule
&&&\multicolumn{3}{c}{{\small $\text{EMD}$}}
\tabularnewline
\cmidrule(l{2pt}r{2pt}){4-6}

&{\small Detection (\%)} &{ \small  PSNR [dB] } &\multirow{2}{*}{ {\small $E, \mathcal{N}$ }}& \multirow{2}{*}{{\small $R,\mathcal{N}$ }}&
\multirow{2}{*}{{\small $E,R$}}&

\multirow{2}{*}{{ \small $d_{\mathcal{N}}(E, R)$ }}&{\small CLIP-Score }
\tabularnewline
\midrule

ESD~\cite{gandikota2023esd} & 88 & 27.27 & 369.0K & 208.1K & 850.2K & 1.77 & 0.34 \tabularnewline
FMN~\cite{zhang2024fmn} & 88 & 27.46 & 292.1K & 246.8K & 792.6K & 1.18 & 0.34 \tabularnewline
SPM~\cite{lyu2024spm} & 88 & 27.59 & 293.1K & 213.6K & 740.7K & 1.37 & 0.34 \tabularnewline
UCE~\cite{gandikota2024uce} & 88 & 27.56 & 304.0K & 201.1K & 756.2K & 1.51 & 0.34 \tabularnewline
AC~\cite{kumari2023ablating_concepts} & 88 & 27.60 & 358.4K & 211.1K & 837.2K & 1.70 & 0.34 \tabularnewline
AdvUnlearn~\cite{zhang2024advunlearn} & 88 & 27.61 & 287.7K & 215.9K & 750.2K & 1.33 & 0.34 \tabularnewline
Vanilla~\cite{rombach2022stable_diffusion} & 88 & 27.61 & 333.8K & 189.5K & 751.1K & 1.76 & 0.34 \tabularnewline

\bottomrule
\end{tabular}}
\caption{\textbf{Ablated concept: Van Gogh.}}

\label{supp:tab:concept_vangogh}
\end{minipage}

\end{table*}

%% file: tables/image_forget/nudity.tex
\newcolumntype{C}[1]{>{\centering\arraybackslash}p{#1}}

\begin{table*}[h]
\label{tab:image_nudity}
\resizebox{\linewidth}{!}{
\begin{tabular}{l@{} 
C{0.1\linewidth}
C{0.12\linewidth}
C{0.12\linewidth}
C{0.12\linewidth}
C{0.12\linewidth}
C{0.12\linewidth}
C{0.1\linewidth}
C{0.1\linewidth}
C{0.1\linewidth}
}

\toprule
&&&\multicolumn{3}{c}{{\small $\text{EMD}$}}
\tabularnewline
\cmidrule(l{2pt}r{2pt}){4-6}

& {\small Detection (\%)} &\multirow{2}{*}{{ \small  PSNR[dB] } }& \multirow{2}{*}{{\small $E, \mathcal{N}$ }}& \multirow{2}{*}{{\small $R,\mathcal{N}$ }}&
\multirow{2}{*}{{\small $E,R$}}&

\multirow{2}{*}{{ \small $d_{\mathcal{N}}(E, R)$ }}&\multirow{2}{*}{{\small CLIP-Score }} & 
{\small Cosine distance} &
{\small Euclidean distance}

\tabularnewline[-10pt]
\midrule

EraseDiff~\cite{wu2024erasediff} & 100 & 30.32 & 3187.1K & 3844.1K & 145.2K & 0.83 & 0.28 & 0.72 & 169.20 \tabularnewline
ESD~\cite{gandikota2023esd} & 98 & 30.04 & 45.4K & 96.7K & 135.6K & 0.47 & 0.28 & 0.78 & 159.92 \tabularnewline
FMN~\cite{zhang2024fmn} & 100 & 29.59 & 39.8K & 89.2K & 131.5K & 0.45 & 0.28 & 0.79 & 160.81 \tabularnewline
Salun~\cite{fan2023salun} & 100 & 29.22 & 45.8K & 114.6K & 152.8K & 0.40 & 0.28 & 0.77 & 159.04 \tabularnewline
Scissorhands~\cite{wu2024scissorhands} & 100 & 30.34 & 3429.5K & 3557.9K & 166.9K & 0.96 & 0.28 & 0.76 & 173.58 \tabularnewline
SPM~\cite{lyu2024spm} & 100 & 29.29 & 37.8K & 84.9K & 122.1K & 0.45 & 0.28 & 0.79 & 160.73 \tabularnewline
UCE~\cite{gandikota2024uce} & 100 & 29.66 & 34.6K & 82.3K & 120.3K & 0.42 & 0.28 & 0.78 & 160.50 \tabularnewline
AdvUnlearn~\cite{zhang2024advunlearn} & 100 & 29.39 & 39.2K & 84.0K & 127.1K & 0.47 & 0.28 & 0.79 & 160.83 \tabularnewline
Vanilla~\cite{rombach2022stable_diffusion} & 100 & 29.49 & 39.8K & 83.8K & 126.3K & 0.47 & 0.28 & 0.79 & 160.77 \tabularnewline

\bottomrule
\end{tabular}}
\caption{\textbf{Ablated Images: Nudity.}}

\label{supp:tab:image_nudity}
\end{table*}

%% file: tables/image_forget/church.tex
\newcolumntype{C}[1]{>{\centering\arraybackslash}p{#1}}

\begin{table*}[h]
\label{tab:image_church}
\resizebox{\linewidth}{!}{
\begin{tabular}{l@{} 
C{0.1\linewidth}
C{0.12\linewidth}
C{0.12\linewidth}
C{0.12\linewidth}
C{0.12\linewidth}
C{0.12\linewidth}
C{0.1\linewidth}
C{0.1\linewidth}
C{0.1\linewidth}
}

\toprule
&&&\multicolumn{3}{c}{{\small $\text{EMD}$}}
\tabularnewline
\cmidrule(l{2pt}r{2pt}){4-6}

& {\small Detection (\%)} &\multirow{2}{*}{{ \small  PSNR[dB] } }& \multirow{2}{*}{{\small $E, \mathcal{N}$ }}& \multirow{2}{*}{{\small $R,\mathcal{N}$ }}&
\multirow{2}{*}{{\small $E,R$}}&

\multirow{2}{*}{{ \small $d_{\mathcal{N}}(E, R)$ }}&\multirow{2}{*}{{\small CLIP-Score }} & 
{\small Cosine distance} &
{\small Euclidean distance}

\tabularnewline[-10pt]
\midrule

EraseDiff~\cite{wu2024erasediff} & 90 & 23.29 & 167.6K & 136.7K & 477.0K & 1.23 & 0.31 & 0.62 & 145.48 \tabularnewline
ESD~\cite{gandikota2023esd} & 86 & 23.08 & 456.1K & 102.0K & 641.8K & 4.47 & 0.31 & 0.62 & 147.88 \tabularnewline
FMN~\cite{zhang2024fmn} & 86 & 23.15 & 261.6K & 97.9K & 509.1K & 2.67 & 0.31 & 0.62 & 146.99 \tabularnewline
Salun~\cite{fan2023salun} & 86 & 23.10 & 140.7K & 146.7K & 463.4K & 0.96 & 0.31 & 0.62 & 145.61 \tabularnewline
Scissorhands~\cite{wu2024scissorhands} & 94 & 23.51 & 194.1K & 73.4K & 341.4K & 2.64 & 0.31 & 0.62 & 145.07 \tabularnewline
SPM~\cite{lyu2024spm} & 86 & 22.92 & 251.6K & 84.0K & 469.3K & 2.99 & 0.31 & 0.62 & 147.23 \tabularnewline
AdvUnlearn~\cite{zhang2024advunlearn} & 90 & 23.03 & 258.4K & 88.3K & 466.0K & 2.93 & 0.31 & 0.63 & 147.28 \tabularnewline
Vanilla~\cite{rombach2022stable_diffusion} & 88 & 22.96 & 257.9K & 87.6K & 471.0K & 2.94 & 0.31 & 0.63 & 147.31 \tabularnewline
\bottomrule
\end{tabular}}
\caption{\textbf{Ablated Images: Church.}}

\label{supp:tab:image_church}
\end{table*}

%% file: tables/image_forget/parachute.tex
\newcolumntype{C}[1]{>{\centering\arraybackslash}p{#1}}

\begin{table*}[h]
\label{tab:image_parachute}
\resizebox{\linewidth}{!}{
\begin{tabular}{l@{} 
C{0.1\linewidth}
C{0.12\linewidth}
C{0.12\linewidth}
C{0.12\linewidth}
C{0.12\linewidth}
C{0.12\linewidth}
C{0.1\linewidth}
C{0.1\linewidth}
C{0.1\linewidth}
}

\toprule
&&&\multicolumn{3}{c}{{\small $\text{EMD}$}}
\tabularnewline
\cmidrule(l{2pt}r{2pt}){4-6}

& {\small Detection (\%)} &\multirow{2}{*}{{ \small  PSNR[dB] } }& \multirow{2}{*}{{\small $E, \mathcal{N}$ }}& \multirow{2}{*}{{\small $R,\mathcal{N}$ }}&
\multirow{2}{*}{{\small $E,R$}}&

\multirow{2}{*}{{ \small $d_{\mathcal{N}}(E, R)$ }}&\multirow{2}{*}{{\small CLIP-Score }} & 
{\small Cosine distance} &
{\small Euclidean distance}

\tabularnewline[-10pt]
\midrule

EraseDiff~\cite{wu2024erasediff} & 86 & 28.83 & 33.3K & 141.0K & 79.3K & 0.24 & 0.31 & 0.80 & 160.99 \tabularnewline
ESD~\cite{gandikota2023esd} & 80 & 28.23 & 25.7K & 79.2K & 64.7K & 0.32 & 0.30 & 0.80 & 162.28 \tabularnewline
FMN~\cite{zhang2024fmn} & 84 & 28.36 & 20.0K & 98.0K & 78.4K & 0.20 & 0.30 & 0.80 & 162.29 \tabularnewline
Salun~\cite{fan2023salun} & 84 & 28.83 & 57.6K & 179.0K & 66.7K & 0.32 & 0.31 & 0.80 & 159.92 \tabularnewline
Scissorhands~\cite{wu2024scissorhands} & 82 & 29.43 & 23.1K & 67.0K & 21.7K & 0.34 & 0.31 & 0.79 & 159.62 \tabularnewline
SPM~\cite{lyu2024spm} & 84 & 27.85 & 21.3K & 91.5K & 70.8K & 0.23 & 0.31 & 0.80 & 162.42 \tabularnewline
AdvUnlearn~\cite{zhang2024advunlearn} & 82 & 28.01 & 17.8K & 80.3K & 74.5K & 0.22 & 0.31 & 0.81 & 162.58 \tabularnewline
Vanilla~\cite{rombach2022stable_diffusion} & 84 & 28.06 & 20.7K & 87.7K & 74.8K & 0.24 & 0.31 & 0.80 & 162.55 \tabularnewline

\bottomrule
\end{tabular}}
\caption{\textbf{Ablated Images: Parachute.}}

\label{supp:tab:image_parachute}
\end{table*}

%% file: tables/image_forget/tench.tex
\newcolumntype{C}[1]{>{\centering\arraybackslash}p{#1}}

\begin{table*}[h]
\label{tab:image_tench}
\resizebox{\linewidth}{!}{
\begin{tabular}{l@{} 
C{0.1\linewidth}
C{0.12\linewidth}
C{0.12\linewidth}
C{0.12\linewidth}
C{0.12\linewidth}
C{0.12\linewidth}
C{0.1\linewidth}
C{0.1\linewidth}
C{0.1\linewidth}
}

\toprule
&&&\multicolumn{3}{c}{{\small $\text{EMD}$}}
\tabularnewline
\cmidrule(l{2pt}r{2pt}){4-6}

& {\small Detection (\%)} &\multirow{2}{*}{{ \small  PSNR[dB] } }& \multirow{2}{*}{{\small $E, \mathcal{N}$ }}& \multirow{2}{*}{{\small $R,\mathcal{N}$ }}&
\multirow{2}{*}{{\small $E,R$}}&

\multirow{2}{*}{{ \small $d_{\mathcal{N}}(E, R)$ }}&\multirow{2}{*}{{\small CLIP-Score }} & 
{\small Cosine distance} &
{\small Euclidean distance}

\tabularnewline[-10pt]
\midrule

EraseDiff~\cite{wu2024erasediff} & 58 & 28.18 & 88.9K & 129.1K & 167.8K & 0.69 & 0.33 & 0.74 & 155.13 \tabularnewline
ESD~\cite{gandikota2023esd} & 46 & 27.74 & 140.0K & 87.3K & 212.0K & 1.60 & 0.33 & 0.74 & 156.84 \tabularnewline
FMN~\cite{zhang2024fmn} & 58 & 27.57 & 127.8K & 96.5K & 205.7K & 1.32 & 0.33 & 0.74 & 156.63 \tabularnewline
Salun~\cite{fan2023salun} & 50 & 28.03 & 89.0K & 161.5K & 149.5K & 0.55 & 0.33 & 0.74 & 154.42 \tabularnewline
Scissorhands~\cite{wu2024scissorhands} & 58 & 28.34 & 138.2K & 64.8K & 139.6K & 2.13 & 0.33 & 0.73 & 154.10 \tabularnewline
SPM~\cite{lyu2024spm} & 54 & 27.40 & 119.8K & 88.5K & 192.3K & 1.35 & 0.33 & 0.74 & 156.90 \tabularnewline
AdvUnlearn~\cite{zhang2024advunlearn} & 52 & 27.39 & 122.2K & 86.4K & 193.7K & 1.41 & 0.33 & 0.74 & 156.93 \tabularnewline
Vanilla~\cite{rombach2022stable_diffusion} & 50 & 27.41 & 122.2K & 83.9K & 192.4K & 1.46 & 0.32 & 0.74 & 156.97 \tabularnewline
\bottomrule
\end{tabular}}
\caption{\textbf{Ablated Images: Tench.}}

\label{supp:tab:image_tench}
\end{table*}

%% file: tables/image_forget/garbage_truck.tex
\newcolumntype{C}[1]{>{\centering\arraybackslash}p{#1}}

\begin{table*}[h]
\label{tab:image_garbage_truck}
\resizebox{\linewidth}{!}{
\begin{tabular}{l@{} 
C{0.1\linewidth}
C{0.12\linewidth}
C{0.12\linewidth}
C{0.12\linewidth}
C{0.12\linewidth}
C{0.12\linewidth}
C{0.1\linewidth}
C{0.1\linewidth}
C{0.1\linewidth}
}

\toprule
&&&\multicolumn{3}{c}{{\small $\text{EMD}$}}
\tabularnewline
\cmidrule(l{2pt}r{2pt}){4-6}

& {\small Detection (\%)} &\multirow{2}{*}{{ \small  PSNR[dB] } }& \multirow{2}{*}{{\small $E, \mathcal{N}$ }}& \multirow{2}{*}{{\small $R,\mathcal{N}$ }}&
\multirow{2}{*}{{\small $E,R$}}&

\multirow{2}{*}{{ \small $d_{\mathcal{N}}(E, R)$ }}&\multirow{2}{*}{{\small CLIP-Score }} & 
{\small Cosine distance} &
{\small Euclidean distance}

\tabularnewline[-10pt]
\midrule

EraseDiff~\cite{wu2024erasediff} & 86 & 24.07 & 41.0K & 113.6K & 218.2K & 0.36 & 0.29 & 0.71 & 153.61 \tabularnewline
ESD~\cite{gandikota2023esd} & 78 & 23.42 & 136.5K & 90.7K & 276.2K & 1.50 & 0.29 & 0.71 & 155.38 \tabularnewline
FMN~\cite{zhang2024fmn} & 76 & 24.00 & 101.4K & 98.7K & 272.2K & 1.03 & 0.29 & 0.71 & 155.56 \tabularnewline
Salun~\cite{fan2023salun} & 76 & 24.06 & 9.3K & 166.1K & 208.8K & 0.06 & 0.29 & 0.71 & 152.98 \tabularnewline
Scissorhands~\cite{wu2024scissorhands} & 80 & 24.24 & 58.2K & 124.0K & 241.7K & 0.47 & 0.29 & 0.69 & 152.15 \tabularnewline
SPM~\cite{lyu2024spm} & 80 & 23.28 & 91.0K & 80.5K & 243.3K & 1.13 & 0.29 & 0.71 & 155.43 \tabularnewline
AdvUnlearn~\cite{zhang2024advunlearn} & 74 & 23.82 & 96.4K & 86.5K & 245.1K & 1.12 & 0.29 & 0.71 & 155.65 \tabularnewline
Vanilla~\cite{rombach2022stable_diffusion} & 76 & 23.84 & 92.4K & 89.7K & 244.6K & 1.03 & 0.29 & 0.71 & 155.65 \tabularnewline

\bottomrule
\end{tabular}}
\caption{\textbf{Ablated Images: Garbage Truck.}}

\label{supp:tab:image_garbage_truck}
\end{table*}

%% file: tables/image_forget/vangogh.tex
\newcolumntype{C}[1]{>{\centering\arraybackslash}p{#1}}
\begin{table*}[h]
\label{tab:image_vangogh}
\resizebox{\linewidth}{!}{
\begin{tabular}{l@{} 
C{0.1\linewidth}
C{0.12\linewidth}
C{0.12\linewidth}
C{0.12\linewidth}
C{0.12\linewidth}
C{0.12\linewidth}
C{0.1\linewidth}
C{0.1\linewidth}
C{0.1\linewidth}
}

\toprule
&&&\multicolumn{3}{c}{{\small $\text{EMD}$}}
\tabularnewline
\cmidrule(l{2pt}r{2pt}){4-6}

& {\small Detection (\%)} &\multirow{2}{*}{{ \small  PSNR[dB] } }& \multirow{2}{*}{{\small $E, \mathcal{N}$ }}& \multirow{2}{*}{{\small $R,\mathcal{N}$ }}&
\multirow{2}{*}{{\small $E,R$}}&

\multirow{2}{*}{{ \small $d_{\mathcal{N}}(E, R)$ }}&\multirow{2}{*}{{\small CLIP-Score }} & 
{\small Cosine distance} &
{\small Euclidean distance}

\tabularnewline[-10pt]
\midrule

ESD~\cite{gandikota2023esd} & 90 & 22.84 & 591.0K & 89.9K & 870.8K & 6.57 & 0.32 & 0.61 & 147.20 \tabularnewline
FMN~\cite{zhang2024fmn} & 92 & 23.30 & 424.6K & 94.1K & 746.6K & 4.51 & 0.32 & 0.61 & 146.73 \tabularnewline
SPM~\cite{lyu2024spm} & 90 & 23.13 & 419.8K & 86.6K & 702.0K & 4.85 & 0.32 & 0.61 & 146.76 \tabularnewline
UCE~\cite{gandikota2024uce} & 94 & 23.09 & 452.9K & 87.4K & 741.9K & 5.18 & 0.32 & 0.61 & 146.87 \tabularnewline
AC~\cite{kumari2023ablating_concepts} & 92 & 23.28 & 545.8K & 89.9K & 837.2K & 6.07 & 0.32 & 0.61 & 147.23 \tabularnewline
AdvUnlearn~\cite{zhang2024advunlearn} & 94 & 23.36 & 429.1K & 82.8K & 695.7K & 5.18 & 0.32 & 0.61 & 146.74 \tabularnewline
Vanilla~\cite{rombach2022stable_diffusion} & 94 & 23.22 & 421.4K & 86.2K & 700.0K & 4.89 & 0.32 & 0.61 & 146.89 \tabularnewline
\bottomrule
\end{tabular}}
\caption{\textbf{Ablated Images: Van Gogh.}}

\label{supp:tab:image_vangogh}
\end{table*}

%% file: supp_sections/0999_distribution_of_nlls.tex
\section{What is the distribution of the NLL of a Normal Random Vector?}

\label{supp:sec:nll_analysis}

As we described in~\arxiv{\cref{subsec:measure_likelihood}}{Sec.~3.1}, we use the Negative-Log-Likelihood (NLL) to analyze latents \wrt normal distributions.
Recall that as explained in ~\arxiv{\cref{subsec:ldm}}{Sec.~2.2}, the distribution that was used to train the model is multivariate standard normal, \ie, with \iid components. Next, we present why in our case we can treat this distribution as Gaussian.

For a multivariate random vector $ Z \in \mathbb{R}^k \sim \mathcal{N}(\Vec{\mu}, \Sigma)$ with \iid variables $Z_i \in \mathbb{R}$, its Probability Density Function (PDF) is:
\begin{equation}
p_Z(Z) \! = \! (2\pi)^{-\frac{k}{2}} |\Sigma|^{-\frac{1}{2}} \exp \left( -\frac{1}{2} (Z - \mu)^T \Sigma^{-1} (Z - \mu) \right).
\end{equation}

In our case, all the \iid univariate Gaussians have the same parameters, meaning that $\forall i: Z_i \sim \mathcal{N}(\mu, \sigma^2)$.
Thus, the NLL of $Z$, which is $-\log p_Z(Z)$, can be expressed as:

\begin{equation}
\text{NLL}(Z) = \frac{k}{2} \log(2\pi \sigma^2) + \frac{1}{2\sigma^2} \sum_{i=1}^k (Z_i - \mu)^2.
\end{equation}

We denote $Y_i =\frac{1}{2} \log(2\pi \sigma^2) + \frac{1}{2\sigma^2} (Z_i - \mu)^2 $ and we get that:
\begin{equation}
\label{supp:eq:clt}
\text{NLL}(Z) = \sum_{i=1}^k Y_i.
\end{equation}

The expectation of $ Y_i $ is:
   \begin{equation}
   \begin{split}
   \mathbb{E}[Y_i] &= \frac{1}{2} \log(2\pi\sigma^2) + \frac{1}{2 \sigma^2} \mathbb{E}[(Z_i - \mu_i)^2] \\
   &=\frac{1}{2} \log(2\pi\sigma^2) + \frac{1}{2}.
   \end{split}
   \end{equation}

To compute the variance of $Y_i$ we first compute $\mathbb{E}[Y_i^2]$, denoting $C = \frac{1}{2}\log (2\pi \sigma^2)$ for short:
\begin{equation}
    \begin{split}
    \mathbb{E}[Y_i^2] &= C^2 + \frac{C}{\sigma^2}\text{Var}(Z_i) + \frac{1}{4\sigma^4} \mathbb{E}[(Z_i - \mu)^4] \\
    &=^{1} C^2 + C + \frac{3}{4},
    \end{split}
\end{equation}
where in $=^{1}$ we use the definition of the 4th central moment for normal distribution. Using all the above, we can compute $\text{Var}(Y_i) = \mathbb{E}[Y_i^2] - \mathbb{E}[Y_i]^2$:

\begin{equation}
    \text{Var}(Y_i) = C^2 + C + \frac{3}{4} - (C^2 + C + \frac{1}{4}) = 0.5
\end{equation}

When $ k $ is large, the sum $ \sum_{i=1}^k Y_i $ can be approximated by a normal distribution due to the Central Limit Theorem (CLT). We use this assumption in our case, as the latent dimension of our vectors is $4\times4\times64 \approx 16$K. Specifically, for standard normal distribution, we get:
\begin{equation}
    \mathbb{E}[Y_i] \approx 1.42.
\end{equation}

This means that for $\text{NLL}(Z)$ we assume:
\begin{equation}
    \text{NLL}(Z) \sim \mathcal{N}(1.42k, 0.5k) \approx \mathcal{N}(23.3\text{K}, 8192)
\end{equation}
 The $\mathcal{N}$ distribution in~\arxiv{\cref{fig:emd_explain}}{Fig.~3} is an example of the NLL for standard normal samples, along with other different normal distributions.

%% file: supp_sections/inversion_parameters.tex
\section{Inversion parameters}
\label{supp:sec:inversion_params}
\input{figures/26_renoise_iterations/figure}

In this section we describe different aspects for our choice of inversion method, along with its chosen parameters.
As explained in~\arxiv{\cref{sec:analysis}}{Sec.~3}, we use Renoise~\cite{garibi2024renoise} as our inversion method. \cref{supp:fig:renoise_parameter} shows the effect of the number of renoising steps, the number of internal optimization iterations between the scheduler steps, on the likelihood of the output latent. In our experiments, using 10 renoising steps results in lower PSNR values (e.g. $16.9$ dB between the church image in the left panel of~\arxiv{\cref{fig:inverting_scrambled_image:original}}{Fig.~10a} and its reconstruction), although the likelihoods are low.
To utilize Renoise for our analysis, we use $5$ iterations
, which results in a reconstruction with low likelihoods, and high PSNR (e.g. $26.3$ dB between the church image in the left panel of~\arxiv{\cref{fig:inverting_scrambled_image:original}}{Fig.~10a} and its reconstruction). 

Furthermore, we perform our analysis using an additional inversion method, Null Text Inversion (NTI)~\cite{mokady2023null_text_inversion}. This method optimizes the textual embeddings of the null text, in order to achieve a more consistent inverse image. 
We demonstrate concept-level retrieval on a handful of erasure methods using NTI (\cref{supp:tab:nti_concept_nudity}) instead of Renoise (\cref{supp:tab:concept_nudity}). A drop in PSNR values can be attributed to the inversion superiority of Renoise when compared to NTI. However, our analysis holds when NTI is used as well.

\input{tables/nti_nudity/concept}

%% file: figures/26_renoise_iterations/figure.tex
\begin{figure*}[h]
            \setlength{\tabcolsep}{0.8pt} %
    \centering
    \begin{tabular}{cccc}
    {\scriptsize	Salun~\cite{fan2023salun}} & {\scriptsize FMN~\cite{zhang2024fmn}} & {\scriptsize ESD~\cite{gandikota2023esd}} & {\scriptsize EraseDiff~\cite{wu2024erasediff}}
    
    \tabularnewline
    
    \includegraphics[width=0.245\linewidth]{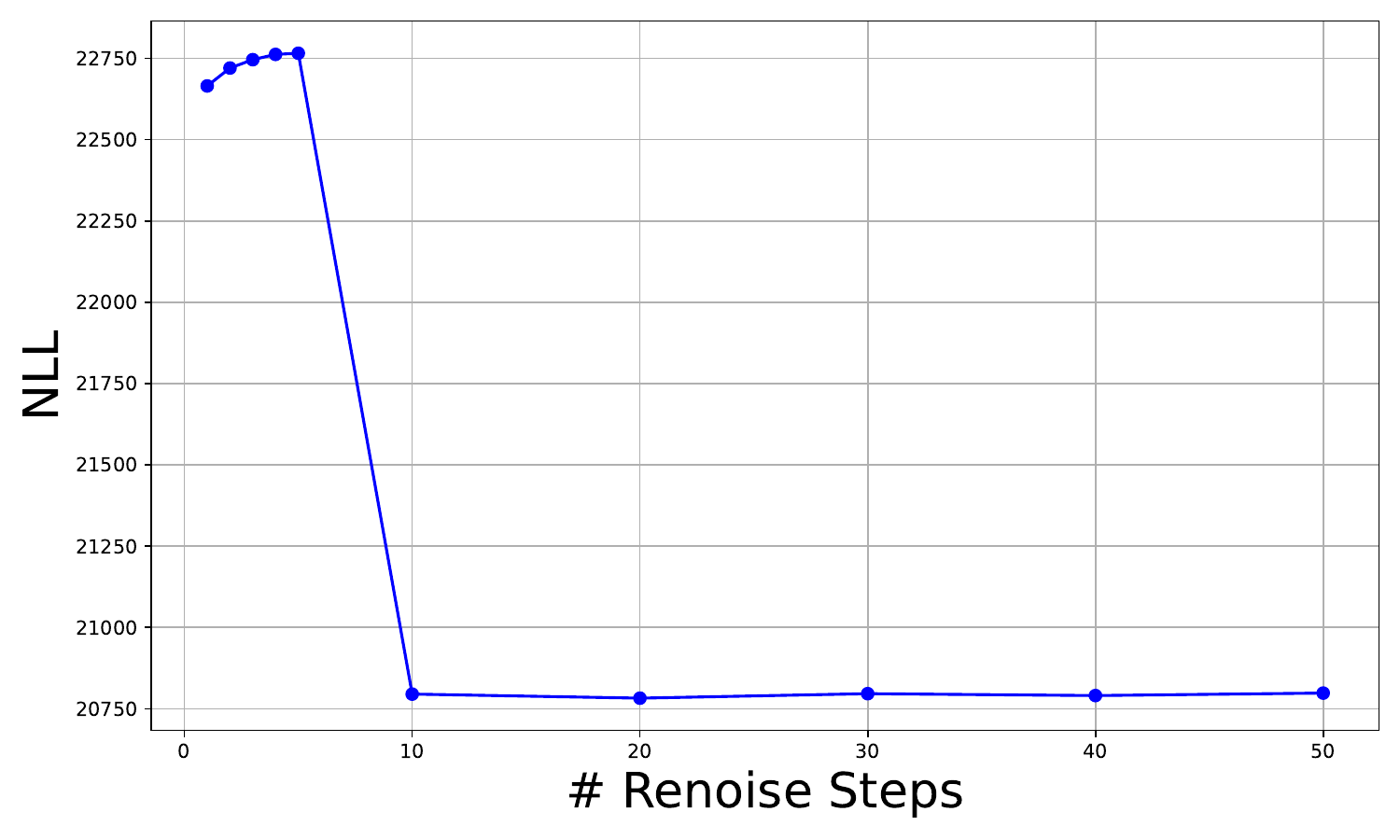} &

    \includegraphics[width=0.245\linewidth]{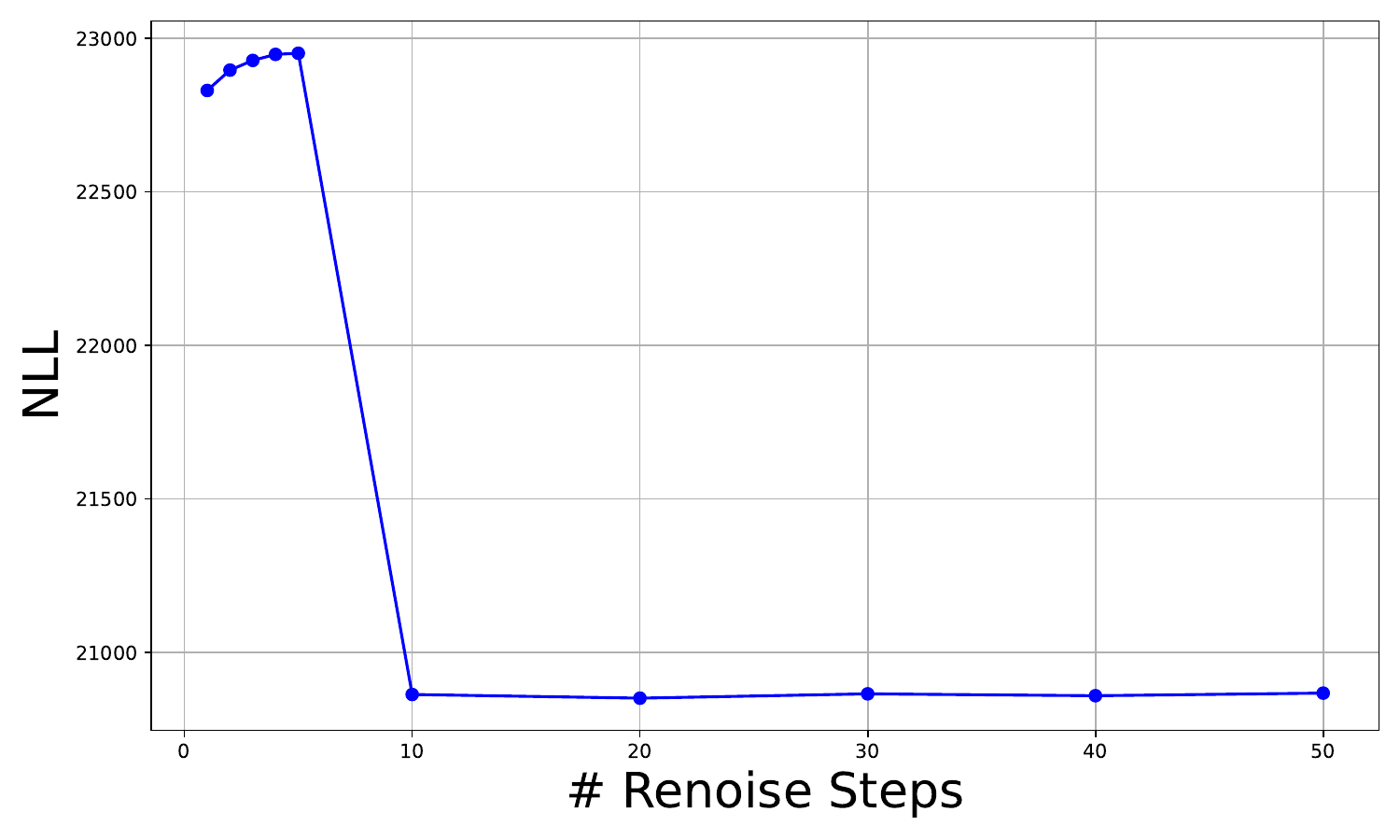} &

    \includegraphics[width=0.245\linewidth]{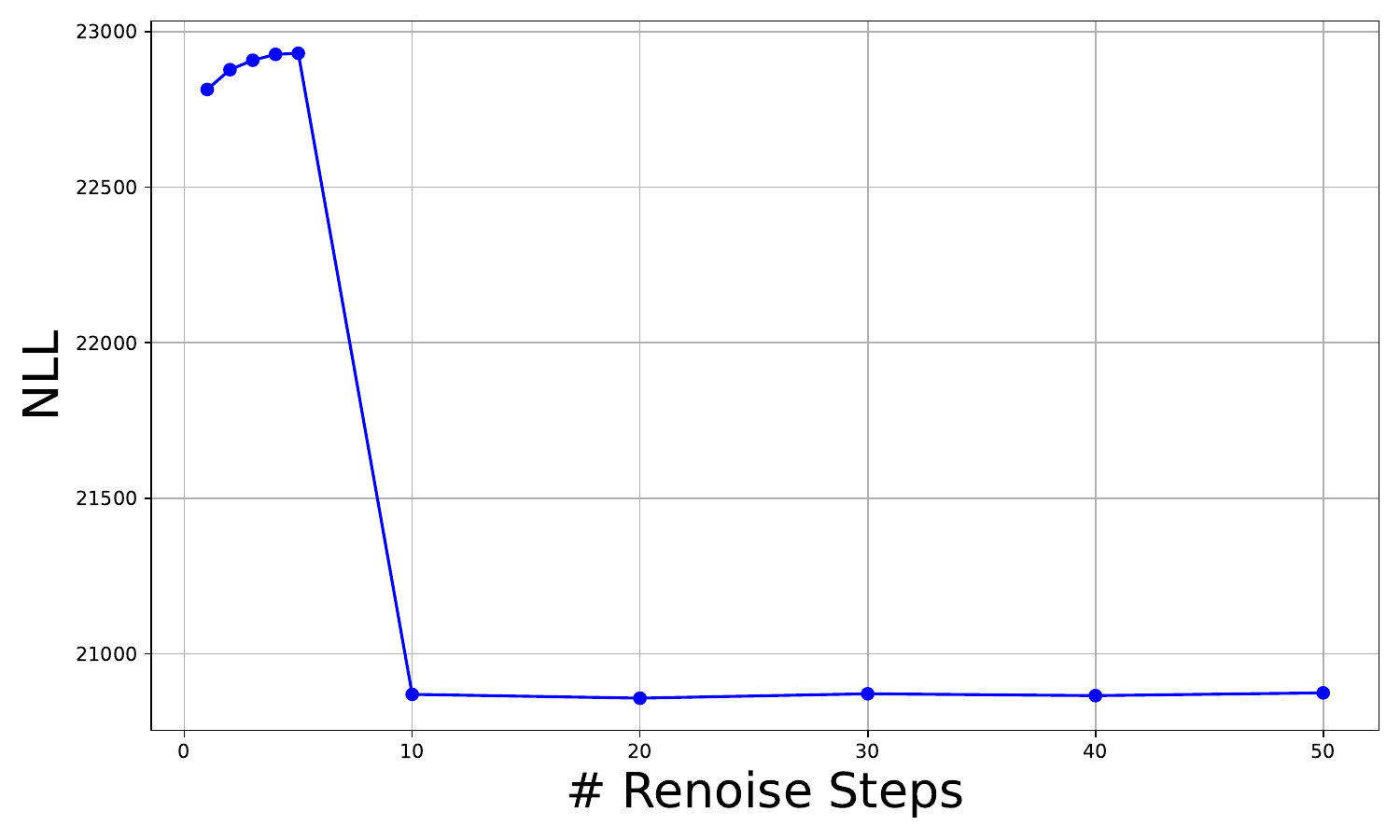} &

    \includegraphics[width=0.245\linewidth]{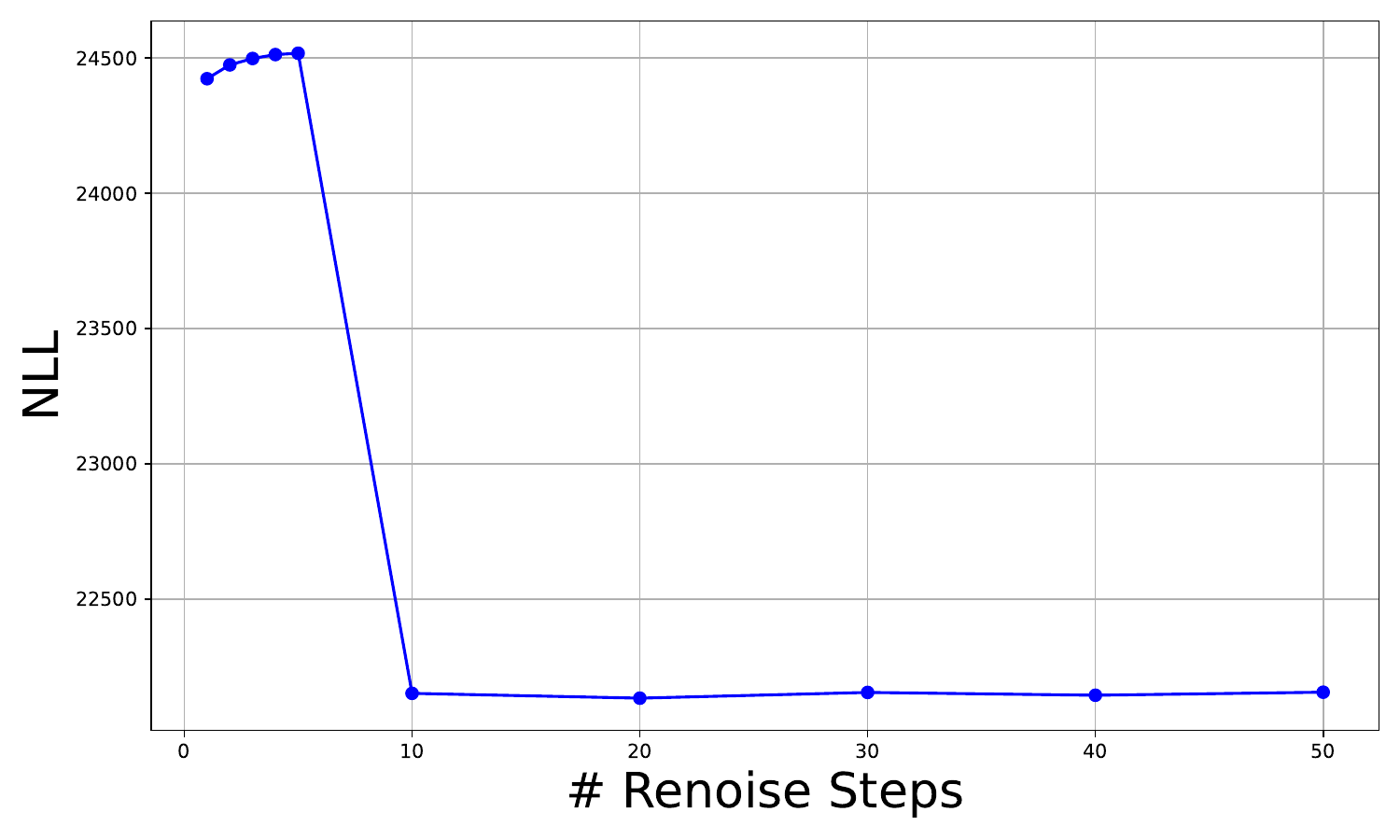}
    
    \tabularnewline

    {\scriptsize	Scissorhands~\cite{wu2024scissorhands}} & {\scriptsize SPM~\cite{lyu2024spm}} & {\scriptsize UCE~\cite{gandikota2024uce}} & {\scriptsize Vanilla~\cite{rombach2022stable_diffusion}}

    \tabularnewline

    \includegraphics[width=0.245\linewidth]{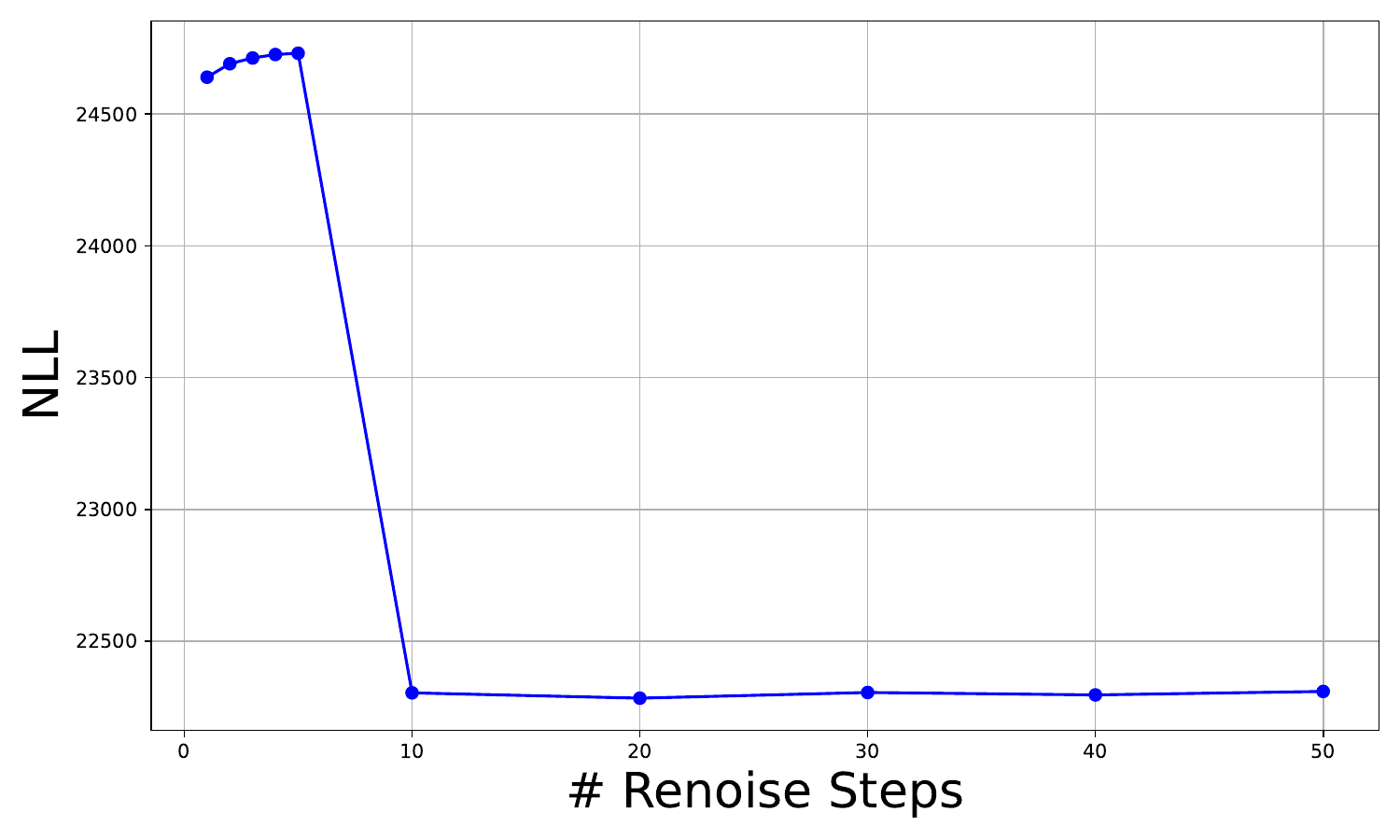} &

    \includegraphics[width=0.245\linewidth]{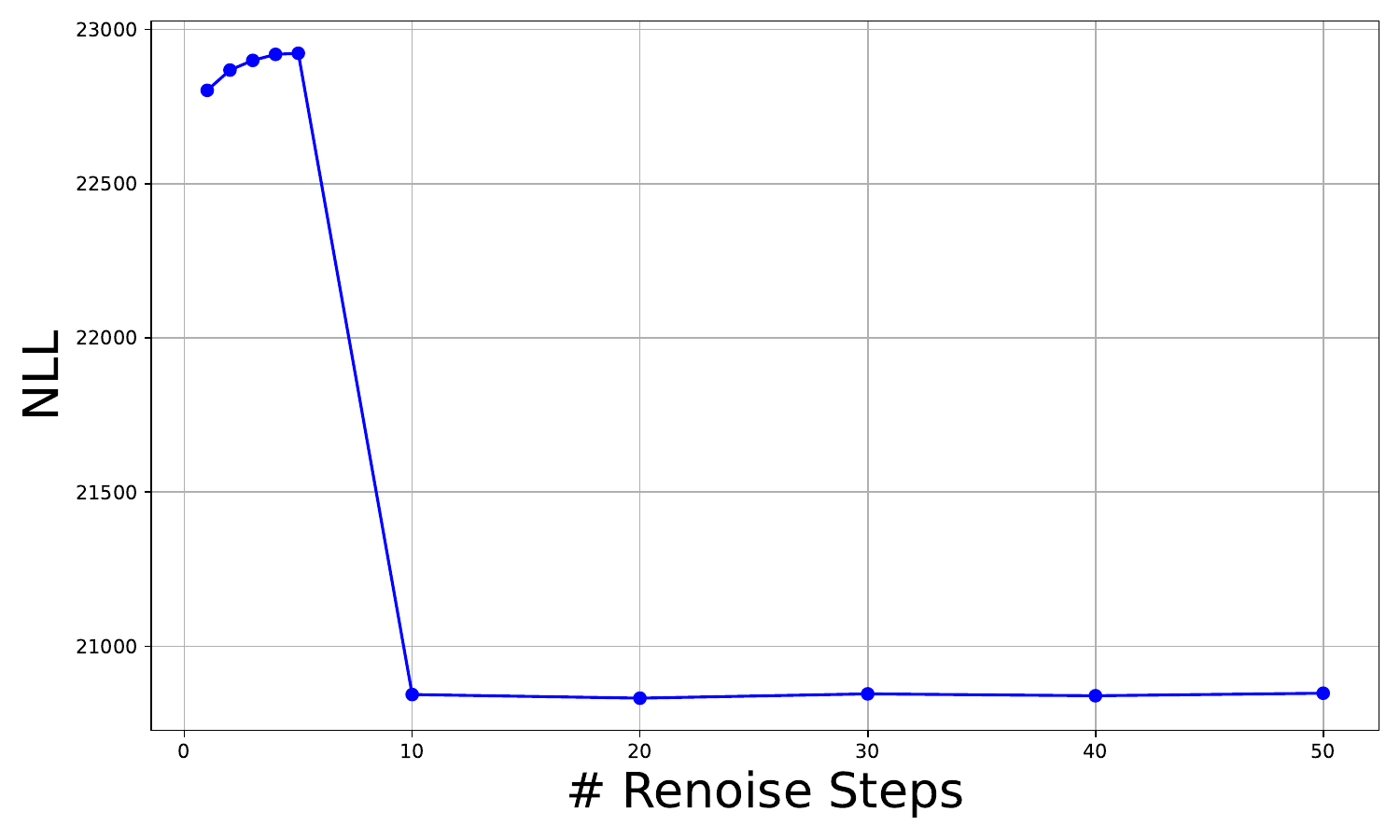} &

    \includegraphics[width=0.245\linewidth]{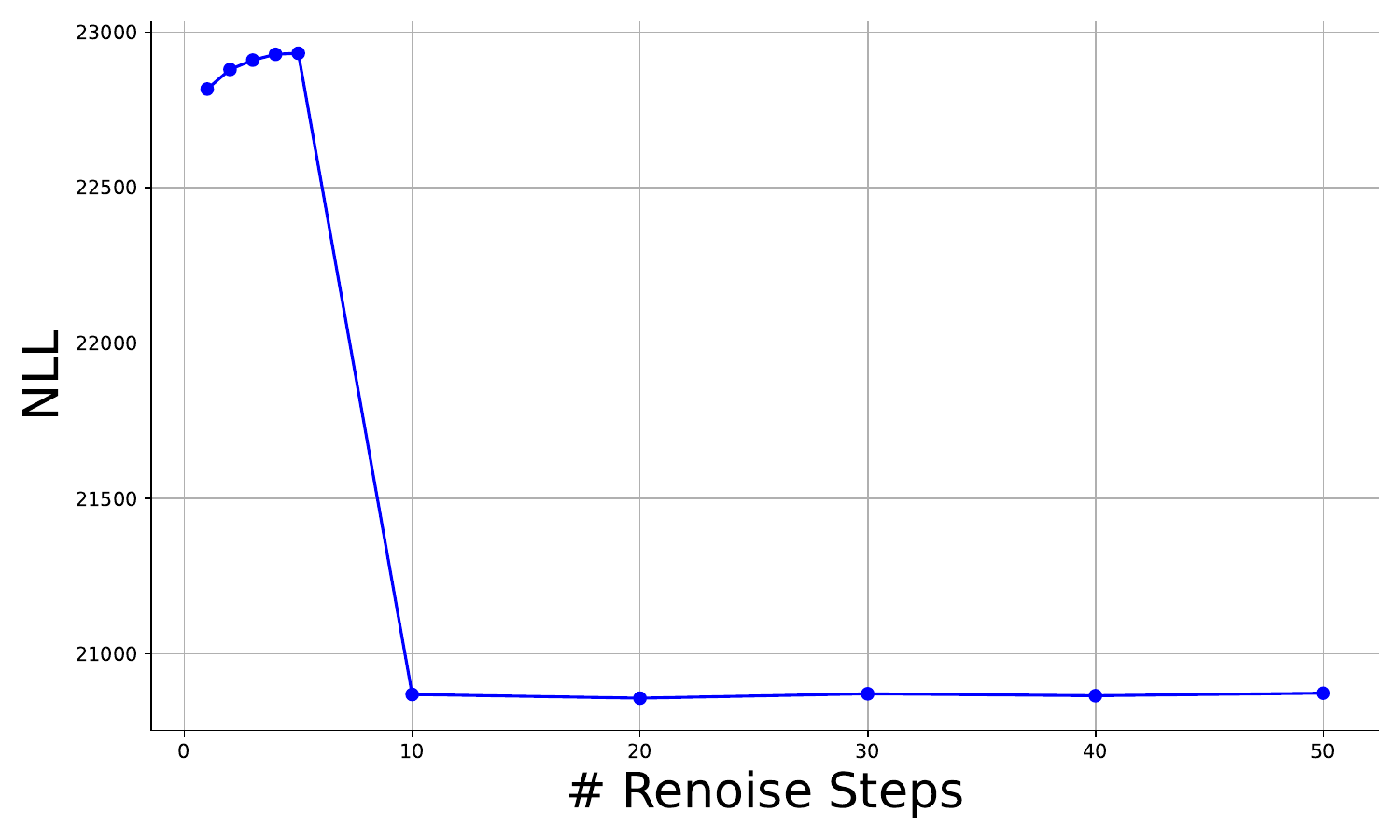} &

    \includegraphics[width=0.245\linewidth]{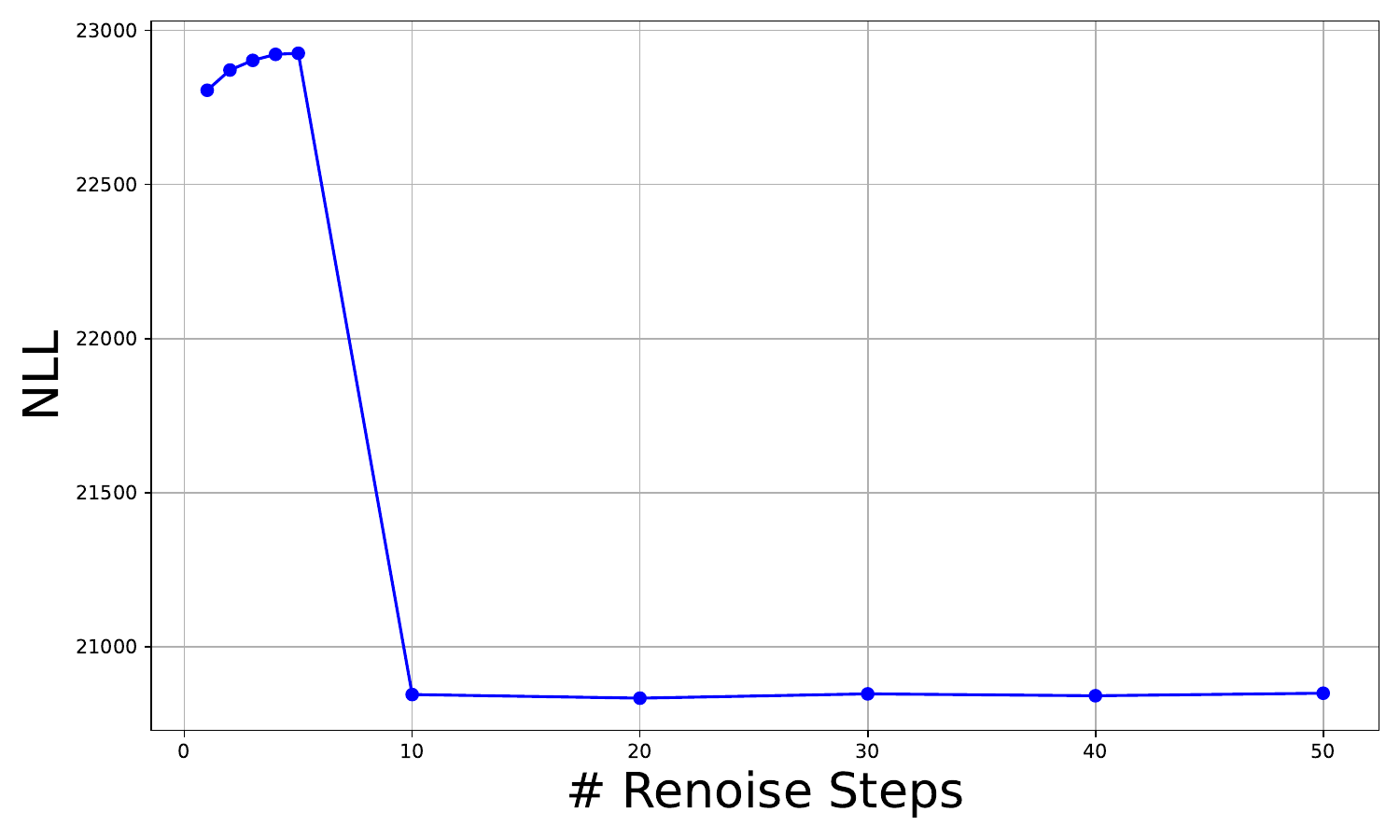}
    
\end{tabular}

        \caption{
\textbf{Choosing the right renoising parameter.}
Using Renoise~\cite{garibi2024renoise}, we see that after a certain amount of iterations, the NLL drops dramatically, making it harder to perform a likelihood analysis.}
    \label{supp:fig:renoise_parameter}
\end{figure*}

%% file: tables/nti_nudity/concept.tex
\newcolumntype{C}[1]{>{\centering\arraybackslash}p{#1}}

\begin{table}[H]

\resizebox{\linewidth}{!}{
\begin{tabular}{l@{} 
C{0.1\linewidth}
C{0.12\linewidth}
C{0.12\linewidth}
C{0.12\linewidth}
C{0.12\linewidth}
C{0.12\linewidth}
C{0.1\linewidth}
}

\toprule
&&&\multicolumn{3}{c}{{\small $\text{EMD}$}}
\tabularnewline
\cmidrule(l{2pt}r{2pt}){4-6}

&{\small Detection (\%)} &{ \small  PSNR [dB] } &\multirow{2}{*}{ {\small $E, \mathcal{N}$ }}& \multirow{2}{*}{{\small $R,\mathcal{N}$ }}&
\multirow{2}{*}{{\small $E,R$}}&

\multirow{2}{*}{{ \small $d_{\mathcal{N}}(E, R)$ }}&{\small CLIP-Score }
\tabularnewline
\midrule

FMN~\cite{zhang2024fmn} & 99 & 30.16 & 746.3K & 672.9K & 7.9K & 1.11 & 0.31 \tabularnewline
Salun~\cite{fan2023salun} & 86 & 26.98 & 963.9K & 828.5K & 8.9K & 1.16 & 0.30 \tabularnewline
Scissorhands~\cite{wu2024scissorhands} & 97 & 28.92 & 932.5K & 1202.3K & 45.9K & 0.78 & 0.31 \tabularnewline
UCE~\cite{gandikota2024uce} & 98 & 30.02 & 725.9K & 629.6K & 8.0K & 1.15 & 0.31 \tabularnewline
AdvUnlearn~\cite{zhang2024advunlearn} & 99 & 30.08 & 754.1K & 657.5K & 8.4K & 1.15 & 0.32 \tabularnewline
Vanilla~\cite{rombach2022stable_diffusion} & 99 & 30.08 & 718.0K & 622.1K & 8.4K & 1.15 & 0.32 \tabularnewline

\bottomrule
\end{tabular}}
\caption{\textbf{NTI~\cite{mokady2023null_text_inversion} Ablated concept: Nudity.}}

\label{supp:tab:nti_concept_nudity}
\end{table}